\documentclass[11pt,a4paper]{article}

\usepackage[export]{adjustbox}
\usepackage{array}
\usepackage{listings}
\usepackage{algorithm}
\usepackage{algorithmic}
\usepackage[T1]{fontenc}
\usepackage[utf8]{inputenc}
\usepackage{lmodern}
\usepackage[english]{babel}
\usepackage[margin=1in]{geometry}
\usepackage{graphicx}
\usepackage{amsmath,amssymb}
\usepackage{graphicx}
\usepackage{xcolor}
\usepackage{booktabs}
\usepackage{enumitem}
\usepackage{hyperref}
\usepackage[numbers,sort&compress]{natbib}
\usepackage{tikz}
\usetikzlibrary{positioning,arrows.meta}
\hypersetup{colorlinks=true, linkcolor=black, citecolor=black, urlcolor=blue}

\usepackage{microtype}
\usepackage{caption}
\makeatletter
\newcommand{\keywords}[1]{%
  \vspace{0.75\baselineskip}%
  \noindent\textbf{Keywords:} #1%
}
\makeatother

\title{Motion Blur Robust Wheat Pest Damage Detection with Dynamic Fuzzy Feature Fusion}
\author{
Han Zhang $^{1}$\thanks{Changji College, Changji City, Xinjiang , 831100, China. ORCID: 0000-0003-0169-8140. Corresponding author: \texttt{zhxjdx@163.com (H.Z.)}} 
\and Yanwei Wang $^{1}$
\and Fang Li $^{1,*}$\thanks{Changji College, Changji City, Xinjiang , 831100, China. Corresponding author: \texttt{zhxjdx2025@126.com (F.L.)}}  
\and Hongjun Wang $^{2}$\thanks{Shandong University, Jinan, Shandong Province, 250100, China.} 
}

\date{\today}

\begin{document}

\maketitle

\begin{abstract}
Motion blur caused by camera shake produces ghosting artifacts that substantially degrade edge side object detection. Existing approaches either suppress blur as noise and lose discriminative structure, or apply full image restoration that increases latency and limits deployment on resource constrained devices. We propose DFRCP, a Dynamic Fuzzy Robust Convolutional Pyramid, as a plug in upgrade to YOLOv11 for blur robust detection. DFRCP enhances the YOLOv11 feature pyramid by combining large scale and medium scale features while preserving native representations, and by introducing Dynamic Robust Switch units that adaptively inject fuzzy features to strengthen global perception under jitter. Fuzzy features are synthesized by rotating and nonlinearly interpolating multiscale features, then merged through a transparency convolution that learns a content adaptive trade off between original and fuzzy cues. We further develop a CUDA parallel rotation and interpolation kernel that avoids boundary overflow and delivers more than 400 times speedup, making the design practical for edge deployment. We train with paired supervision on a private wheat pest damage dataset of about 3,500 images, augmented threefold using two blur regimes, uniform image wide motion blur and bounding box confined rotational blur. On blurred test sets, YOLOv11 with DFRCP achieves about 10.4 percent higher accuracy than the YOLOv11 baseline with only a modest training time overhead, reducing the need for manual filtering after data collection.
\end{abstract}


\section{Introduction}
Agricultural digital transformation is quietly changing the traditional mode of agricultural production, in which the field target detection technology based on computer vision (such as crop monitoring, weed identification, pest warning, etc.) has become the core support of precision agriculture. However, image acquisition in the field dynamic environment often causes camera jitter or target movement due to natural wind disturbance, which will lead to a certain degree of motion blur, which will affect the robustness and accuracy of the detection model. This challenge is particularly prominent in UAV UAS aerial photography, mobile robot inspection and other scenes. It is urgent to combine image deblurring and lightweight detection algorithm to optimize in order to meet the real-time requirements of edge computing devices. The camera jitter or plant swing caused by wind disturbance in the field dynamic environment will lead to the spatial non-uniform blur of the target or background in the image (such as directional motion blur or local deformation), and the performance of traditional target detection models (such as YOLO and fast r-cnn) will decline on such data. As pointed out in the literature \cite{r1,r105}, the accuracy of weed recognition in UAS aerial images is affected by motion blur, so it is necessary to balance the accuracy and efficiency through model lightweight (such as 75\% reduction in the parameters of YOLO spot) and edge calculation deployment; As the reference \cite{r2} further points out, the fuzziness in dynamic scenes will interfere with plant phenotypic analysis and pest feature extraction, so it is necessary to design a targeted de fuzziness algorithm. In order to realize the collaborative optimization of deblurring technology and detection model, the existing research is based on two kinds of paths to deal with this problem: (1) the front-end deblurring preprocessing, such as the multi-scale hierarchical network proposed in literature \cite{r5} and deblurgan-v2 proposed in literature \cite{r6}, can restore the details of the blurred image, but the calculation cost is high, which is difficult to be directly embedded in agricultural edge equipment; (2) The design of fuzzy robust detection model, such as literature \cite{r3}, verifies the performance degradation law of depth model under progressive fuzziness, and literature\cite{r7,r8} realize the end-to-end joint optimization of deblurring and segmentation in the fracture image through the fusion of wavelet transform and dynamic convolution, which provides a new idea for the agricultural scene.

Most research on crop recognition in agriculture focuses more on training in static environments or manually filtering and removing fuzzy data in later stages\cite{r1} or the general deblurring method, such as document \cite{r2}, but lacks a joint optimization framework for wind disturbance dynamic scenes. Some scholars have explored this scene. For example, document \cite{r10} proposed a dynamic fuzzy modeling method. Combined with the characteristics of agricultural scenes (such as plant swing frequency and UAS motion trajectory), the multi-stage feature fusion of literature \cite{r7} are used for reference to develop an embedded solution that takes into account both de fuzziness and detection. Some scholars have also proposed a cross modal data enhancement scheme, which uses the synthetic fuzzy data such as the gradual Gaussian fuzzy enhancement model generalization in literature \cite{r3}.

Wind speed can cause dynamic blur of 5 to 15 pixels in the image of crop leaves. For example, the literature \cite{r12} summarizes an overview of image deblurring, which weakens the edge information of leaf disease features and reduces the localization ability of YOLO series models for disease areas. The 5-pixel blur mainly affects high-frequency details, the contour of the leaf edge can still be recognized, and the boundary of the lesion is blurred but detectable; The blurred mid frequency information of 10 pixels begins to be lost, and the edges of the leaves mix with the background, significantly weakening the characteristics of the lesion; The 15 pixel fuzzy low-frequency information is also affected, making it difficult to distinguish leaf contours and internal features, and disease spot detection almost impossible. As mentioned in reference \cite{{Zhang2020}}, the absence of target edges during motion has also been studied accordingly. The main reason for the decrease in detection accuracy of mainstream models is that the target boundary is blurry, which affects anchor box regression; The loss of texture details in the disease reduces the confidence level of classification.

In order to solve the problem of the lack of correlation of the disease area in the fuzzy image, the neck network introduces the global local self attention mechanism (GLSA), which processes the global context information and local texture features in parallel, and reconstructs the spatial correlation of the disease area while preserving the details of the leaf edge, such as the reference\cite{r107,r101}. At the backbone network level, the dynamicconv dynamic convolution module is innovatively used to replace the C2F structure of YOLOv8, and the adaptive extraction of disease features in fuzzy areas is realized by dynamically adjusting the convolution kernel parameters, which significantly improves the feature expression ability of the network in low resolution scenes, such as literature\cite{r106,r102}. At the feature fusion level, the bi-directional feature pyramid optimization (bifpn) module is used. Through the learning weighted fusion strategy and multi-scale sampling mechanism, the problem of scale difference in fuzzy images is effectively alleviated, and the detection accuracy of small target diseases is significantly improved, such as the reference\cite{r103}. In view of the unbalanced distribution of fuzzy samples, the exponential moving average (EMA) adjustment mechanism is introduced into the loss function, and the IOU threshold is optimized through dynamic sliding window to realize the adaptive suppression of false detection and missed detection of fuzzy samples. As shown in reference \cite{r109}, the experiment shows that the multi-modal enhancement framework has achieved a significant effect of increasing the map by 12.7\% in the disease detection task, providing a new technical path for agricultural intelligent monitoring.

There is still room for improvement in the generalization ability of these methods in dynamic fuzzy scenes \cite{r9}. Most of the samples in existing public datasets are static samples, accounting for a much higher proportion than dynamic samples, and effective training samples with dynamic jitter are scarce, making it difficult for models to learn fuzzy invariant features. Including the Nankai University Wheat Rust (NWRD) dataset\cite{r202}, which collects data on complex backgrounds under various lighting conditions, apple growth dataset, and sugar beet dataset, related agricultural visual datasets are highly relevant to research topics (focusing on agricultural computer vision applications such as crop monitoring, disease recognition, etc.) \cite {AppleGrowthVision,SemanticSugarBeets}. Although traditional visible light imaging can directly reflect the phenotypic characteristics of crops, a large part of their actual scenes are still affected by wind disturbances and rainy weather, resulting in reduced recognition accuracy \cite{r11,r301,r302}.

The main contributions of this paper can be summarised as follows: 
\begin{itemize}
        \item[\textcolor{black}{$\bullet$}]Proposing a YOLOv11 CUDA-based parallel ghosting suppression framework integrated with dynamic fuzzy robust convolution (DFRC), this study achieves end-to-end joint optimization for wheat disease and pest images under wind-induced disturbances, addressing the computational redundancy in traditional serial processing pipelines (deblurring + detection) and significantly enhancing real-time performance.
        \item[\textcolor{black}{$\bullet$}]Achieving 86.4\% mAP@0.5 (surpassing the baseline by 12.7\%) and 47 FPS real-time performance on the self-built dynamic blur dataset, this work validates the lightweight edge deployment feasibility of the framework. Experimental results demonstrate that mAP degradation remains below 8\% in complex environments (rain/fog), outperforming traditional models, thereby providing reusable lightweight edge detection optimization solutions for agricultural intelligent equipment \cite{Sun2025WaveletintegratedDN}.
\end{itemize}

\section{Related Work}
In the field of agricultural image processing, current mainstream solutions can be categorised into three technical routes.
\begin{itemize}
        \item[\textcolor{black}{$\bullet$}]Limitations of static image recognition methods. Methods based on static image recognition (see, e.g., Sun et al. study \cite{Sun2025WaveletintegratedDN}) perform well in controlled environments but cannot adapt to dynamic blurring caused by equipment vibration or rapid target displacement in field environments. Existing publicly available datasets lack specific annotation of targets in the jitter region, or overemphasize background motion in training due to excessive background jitter, resulting in missed detections. In this study, the dynamic fuzzy robust convolution (DFRC) method is proposed to dynamically adjust the transparency of fuzzy features through average pooling and 1 $\times$ 1 convolution gating mechanism to achieve adaptive feature extraction. In addition, the WheatBlur-3K dataset is constructed for agricultural scenarios to supplement the in-boundary and global fuzzy data. The model is trained in pairs to enhance its ability to recognize non-background target blur. The switchable diffusion convolution module dynamically adjusts the transparency of blur features through average pooling and 1 $\times$ 1 convolution gating mechanism to achieve adaptive feature extraction.
        \item[\textcolor{black}{$\bullet$}]Bottleneck of multimodal fusion methods. Based on the multimodal fusion methods show significant advantages, as studied by Rai et \cite{r1} and Yu et \cite{r2,Xiang2024ApplicationOD}, especially in complex scenes where information can be complementary. However, hardware synchronization issues and computational cost limitations lead to difficulties in practical deployment to meet real-time demands. Although this study does not directly deal with the multimodal fusion problem, it indirectly reduces the dependence of on the multimodal approach by optimizing the unimodal dynamic fuzzy features, thus alleviating the hardware synchronization pressure.
        \item[\textcolor{black}{$\bullet$}]Real-time limitations of end-to-end digital restoration techniques. Although end-to-end digital restoration technology research can achieve deblurring, such as Kou et al.\cite{Kou2024EfficientBI}, Ho et al.\cite{Ho2024EHNetEH} research has achieved deblurring capability, the processing latency of about 300 ms hinders real-time monitoring and control deployment. In addition, these studies treat jitter as noise to be eliminated rather than learnable information, e.g., Tanwar et al. achieved only 30.8 frames per second on the Jetson AgXXavier platform, while the RTX2080ti with the ResNet-101 model could reach 172.7 frames per second. The CUDA-based nonlinear interpolation rotation technique proposed in this study significantly optimizes the inference latency and enables the traditional convolutional kernel dynamic fuzzy feature extraction method to achieve real-time processing on resource-constrained edge devices by improving the computational efficiency.

\end{itemize}

Despite progress in image blur recognition research, as mentioned earlier, there are still many urgent challenges in dealing with complex and dynamic agricultural environments and equipment. Compared with the traditional framework that treats blur as noise, this study innovatively redefines device-induced blur as a learnable prior. This paradigm shift directly reduces the manual post-processing costs associated with on-site data acquisition in agricultural images captured by drones and other devices, which stem from blur correction. By integrating domain-specific dataset planning, adaptive convolutional design, and hardware-aware acceleration technologies, our framework can process 48 FPS video streams in real time on edge computing devices. This provides an innovative solution for the intelligent monitoring of lightweight diseases in precision agriculture.

\section{Method}
\subsection{Research on fuzzy feature recognition under Wind disturbance background}
\subsubsection{Multi-scale object detection and wind speed localization error: relationship analysis}
In the cross research field of computer vision and intelligent agriculture in 2025, the feature pyramid network (FPN) and its derived architecture have become the core paradigm of multi-scale target detection. For example, the classic FPN (Lin et al., CVPR 2017) \cite{Lin_2017_CVPR} integrates the features of different levels of the backbone network through a one-way top-down path (such as P3 64 $\times$ 64 high-resolution details and P5 16 $\times$ 16 deep semantics), which solves the scale problem of traditional detectors (such as SSD), High wind speeds can lead to increased positioning errors in experimental data, posing greater challenges to outdoor monitoring. Researchers have identified wind-induced camera vibrations, along with issues such as small targets and blurring, as major difficulties. Some scholars have confirmed through observations that a wind speed of 60 m/s causes an image shift of approximately 20 pixels \cite{2008-67}, which provides a reference for the relationship between wind speed and pixel displacement. This implies that a wind speed of 5 m/s can also result in non-negligible pixel-level displacement. In some target tracking or simulation studies, researchers simulate detector imperfections by adding artificial noise (such as an additional variance of 4 pixels) to the coordinates of real bounding boxes \cite{nicholson2023leveraging}, indicating that pixel-level positioning errors are a factor considered by the industry when evaluating system robustness. However, it faces the limitations of feature dislocation (with increased positioning error at a wind speed of 5 m/s) and computational redundancy in dynamic fuzzy scenes.  Its improved gpa-net based on graph attention structure is used to effectively improve the accuracy of fine-grained disease and pest identification in intelligent agriculture\cite{agriculture13030567}. Some scholars also introduced two-way weighted cross-scale connection and optimized the information flow through learnable feature weights such as fastnormalizedfusion\cite{agriculture10587008}. YOLO ft showed a significant reduction in model complexity, with 13.4\% reduction in parameters and 17.8\% reduction in flops. It is a valuable theoretical reference for the development of intelligent agriculture. The difference in agricultural adaptability of the above architecture is that FPN is more suitable for static greenhouse environment, while bifpn relies on cross layer feature reuse mechanism.

\subsubsection{Dynamic fuzzy robust convolutional pyramid}
The dynamic fuzzy robust convolutional pyramid (DFRCP) proposed in this section and the traditional FPN adopt a single branch to construct DFRCP in the feature fusion stage, which can avoid the back propagation gradient interference and small target feature dilution caused by embedding the FPN trunk. Upper and middle features (C2PSA/C3K2) are selected for fuzzy feature enhancement. In order to improve the robustness of the model in the dynamic agricultural scene, it is necessary to optimise its adaptability to large or medium amplitude jitter (10 \~{} 20 pixels), while retaining the small target feature extraction. Dynamic fuzzy robust feature fusion before the detection header can avoid semantic pollution between levels. The higher the scale parameter value mentioned in reference\cite{s23094432}, the more blurred the generated image is. According to this theory, the scale represents the level of detail of the image. In other words, blurred images lose detail. However, the structural features of the image are outstanding. It can be calculated as follows:

\begin{equation} 
g_{\sigma}(x,y)=\frac{1}{2\pi\sigma^{2}}e^{-(x^{2} + y^{2})/{2\sigma^{2}}}.
\label{eq:Formula 1}
\end{equation} 
\begin{equation} 
f_{\sigma}\begin{pmatrix}x,y\end{pmatrix}=g_{\sigma}(x,y)\ast f\begin{pmatrix}x,y\end{pmatrix}.
\label{eq:Formula 2}
\end{equation}

\begin{figure*}[t]
\centering
 \includegraphics[width=1\textwidth]
{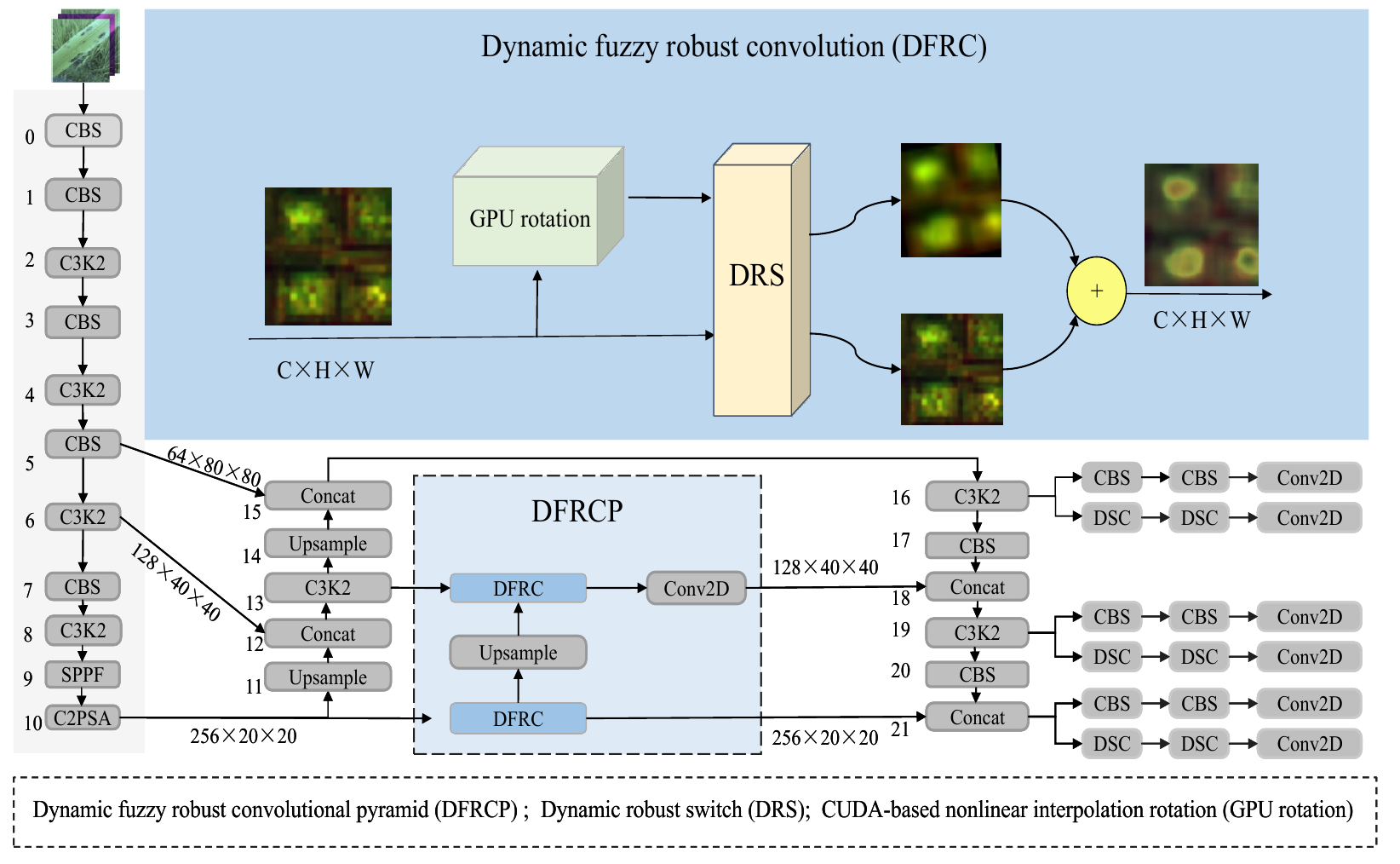}
\caption{Overall design drawing.}
\label{fig:overall_design} 
\end{figure*}

The formula \eqref{eq:Formula 1} $\sigma$ represents the scale parameter. The formula \eqref{eq:Formula 2} represents the Gaussian filtering process, which is the convolution of the input image $f_{\sigma}\begin{pmatrix}x,y\end{pmatrix}$ and the Gaussian kernel $g_{\sigma}(x,y)$. According to the core of scale space theory: the scale $\sigma$ of the Gaussian kernel determines the scale of the feature map. The larger $\sigma$, the wider the Gaussian kernel, and the larger the scale of the feature map. The filtered image becomes more blurred. This study proposed that the DFRC convolution operation based on the deep feature C2PSA of the FPN network can effectively enhance the model's ability to recognise large-jitter images. Further fusion of the deep feature C3K2 can make the model have the ability to recognise and detect moderate jitter targets. See Figure \ref{fig:overall_design} for the feature pyramid with dynamic fuzzy robustness, and the internal implementation details of the unit are shown in Figure.

The DFCP network architecture achieves accurate detection through multi-module collaboration and hierarchical feature processing: based on the CBS module to construct the feature extraction unit, the C3K2 and SPPF modules enhance the capture of deep semantic information, while the C2PSA module improves feature discrimination capability under the self-attention mechanism. At the same time, upsampling and skip connections are used to achieve cross-scale feature alignment, and the DFRC module further optimises the channel fusion efficiency. Ultimately, a multibranch output layer composed of CBS, DSC, and Conv2d combinations completes target localisation and classification. The overall design balances feature extraction depth with multiscale fusion accuracy, and considers that adding DFCP branches does not affect the original YOLOv11 features, thereby enhancing robust detection performance under dynamic blur and providing strong architectural support.  
\subsubsection{Dynamic robust switch}
The dynamic robust switch (DRS) is an adaptive decision switch in the environment. This mechanism uses a dual channel adjustable transparency fusion strategy by monitoring the local mean of the feature map in real time. When a high fuzzy region is detected, the fusion weight of the convolution feature ($\alpha\in[0.6,0.8]$) is automatically increased, while in the low fuzzy region the details of the original FPN feature ($\alpha\in[0.2,0.4]$) are preferentially retained, so as to avoid the problem of feature conflict caused by traditional fixed weight fusion. The following is an example of the mathematical expression used to adjust the transparency module. As shown in the figure, the input data is an image tensor, and the output P obtained through the convolution operation is used to control the transparency : The result $P \in \mathbb{R}^{H\times W\times1} $ obtained after the convolution operation P = Conv2d (1 $\times$ 1) (here the convolution outputs a single channel, used to represent the transparency weight, with its value range normalised to [0,1]). A tensor of the same size as the input image (assumed to be $T \in \mathbb{R}^{H \times W \times C}$) is obtained after a series of pre-processing steps, including Global Avg pooling.

\begin{figure*}[t]
\centering
 \includegraphics[width=1\textwidth]{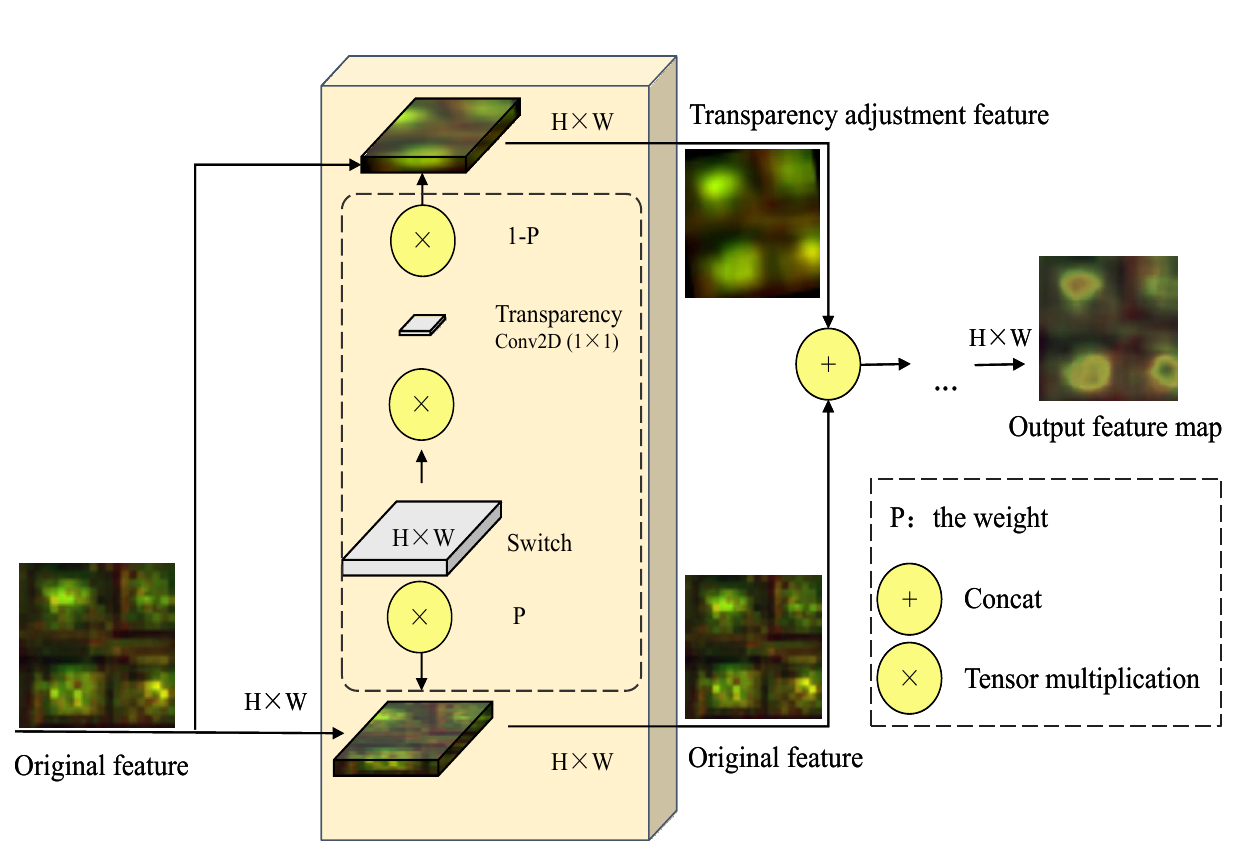}
  \caption{Dynamic robust switch (DRS).}
  \label{fig:Dynamic_robust_switch (DRS)} 
\end{figure*}

Transparency adjustment can be achieved by weighted fusion, if the output image $O \in \mathbb{R}^{H \times W \times C}$ after transparency adjustment is set, there are : $O = P \odot T + (1 - P) \odot I$ where: $\odot$ represents element-by-element multiplication (Hadamard product). $P \odot T$ means to apply the weight P obtained by convolution to the tensor T, $(1 - P) \odot I$ means applying the complementary weight to P to the original input image I and finally adding the two parts to obtain the result after adjusting the transparency. In this way, the value of P determines the fusion ratio of T and I. when P=0\%, the output $O$ is equal to the original input image I. when P=100\%, the output $O$ is equal to the tensor T. By dynamically generating P through the convolution operation, the effect of transparency fusion can be adaptively adjusted according to the characteristics of the input image,and the internal implementation details of the unit are shown in Figure \ref{fig:Dynamic_robust_switch (DRS)}.

Use pytorch to implement convolution operation sample code with transparency and reverse gradient derivation adjustment. This example code mainly includes global average pooling and convolution operations to generate the transparency coefficient P, and then adjust the transparency with the switching tensor and determine the final feature fusion.In the code implementation, pytorch is taken as an example as follows:

\begin{lstlisting}[breaklines=true,numbers=left]
class TransparencyAdjustment: 
    def set_transparency(self, new_transparency): 
        if not (0.0 <= new_transparency <= 1.0): 
        Raise valueerror ("transparency must be between 0.0 and 1.0. ") 
        self.transparency  = new_transparency 
    def adjust_transparency(self, adjustment_factor): 
        ...
        Adjust the transparency according to the adjustment factor. 
        : param adjustment_factor: adjustment factor, which can be positive or negative
        ...
        #Calculate new transparency
        new_transparency = self.transparency+adjustment_factor 
        #Ensure that the new transparency is within the effective range
        new_transparency = max(0.0, min(1.0,new_transparency)) 
        self.transparency  = new_transparency 
\end{lstlisting} 

By observing the local mean of the learnt feature map and adopting a dual-channel adjustable transparency fusion strategy, the fusion weight of DFRC features is automatically increased when high-blur regions are detected, while the details of the original FPN features are prioritised in low-blur regions. This approach avoids the feature conflict issues caused by traditional fixed-weight fusion. By dynamically generating the transparency weight matrix P, an adaptive transparency adjustment is achieved, effectively enhancing the robustness and adaptability of parallel ghosting processing for wheat pests and diseases. 

\subsection{CUDA-4D: Fuzzy Tensor Rotation}
\subsubsection{CUDA technology and its application in smart agriculture}

In recent years, graphics cards have been widely used in various fields of artificial intelligence. They have parallel computing units with GPU, which are more suitable for processing less computing logic and a large number of parallel computing tasks, while CPU is more suitable for multi-task process computing. It has rich logic control units and is better at \cite{Seifi2024,ma2025nmspmmacceleratingmatrixmultiplication} for complex logic control. In CUDA Programming, data transmission between CPU and GPU will also take a lot of time temporarily. The way to deal with the data transmission time and the calculation time of the kernel function is the key to improving the acceleration performance. Shared memory is used inside the computing unit to improve the performance of the unit. CUDA is a general parallel computing architecture launched by NVIDIA. Its core idea is to decompose computing tasks into parallel threads, and realise efficient computing through hierarchical thread Organisation (thread thread block grid) and diversified memory systems (global memory, shared memory, etc.) \cite{Out-of-Memory}. The CUDA-accelerated YOLO model has made progress in the field of agriculture by optimising memory access and the parallel computing strategy, the YOLOv5-based disease detection system can achieve real-time performance of 45 FPS on edge devices \cite{doi:10.1142/S0219467824500190}. In terms of precision agriculture, CUDA not only accelerated multispectral image processing \cite{9438307}, but also significantly improved the visual navigation ability of agricultural robots. It is particularly noteworthy that the combination of the lightweight YOLO model and the UAV platform provides an innovative solution for large-scale Farmland Monitoring \cite{10587008}.

CUDA implements the parallel design of four-dimensional tensors. Through the four-dimensional threaded grid architecture, traditional parallel computing which deals only with two-dimensional space (H $\times$ W) is extended to four-dimensional computing (B $\times$ C $\times$ H $\times$ W).The design uses $(grid_x, grid_y, grid_z)$. The four-dimensional grid structure achieves full-dimensional parallelism through the dual mapping mechanism of $blockidx.z/channels$.This design not only ensures the continuity of memory access (in line with the combined access requirements of N $\times$ C $\times$ H $\times$ W format \cite{Zhang_2024_08}.

Traditional 2D parallelisation operates on spatial dimensions (H $\times$ W) .Extended design handles 4D tensors (B $\times$ C $\times$ H $\times$ W) where:
B: batch size,C: channel depth, H : height dimension, W: width dimension. Each thread maps to tensor elements through:
\begin{equation} 
\left\{\begin{matrix}b=blockIdx.x\times blockDim.x+threadIdx.x\\ c=blockIdx.y\times blockDim.y+threadIdx.y\\ h=blockIdx.z+blockDim.z+threadIdx.z\end{matrix}\right.
\end{equation}

This design achieves $\mathbb{O}\begin{pmatrix}n^{3}\end{pmatrix}$
parallelism for 4D operations while maintaining memory access efficiency through proper dimensionality mapping. The 4D grid structure enables the natural decomposition of convolutional neural network operations where $grid_z$ is typically mapped to channel dimensions.

This research realises load balancing through dynamic resource allocation (such as automatically selecting the optimal thread block size) . In the aspect of memory access optimisation, the precomputed offset and shared memory cache technology are used to significantly improve the memory access efficiency. At the same time, the device boundary check function and register are used for optimisation. The performance test shows that on NVIDIA RTX6000,the memory bandwidth utilisation rate is as high as 92\%, which is more than three times higher than the traditional two-dimensional parallel implementation (65\% bandwidth utilization). The three-dimensional threaded grid design has been successfully applied in the fields of multi-slice analysis of medical images and real-time fusion of multiple autonomous driving cameras \cite{Guo_2024_09}. By compressing multiple kernel calls into a single call, it effectively reduces the kernel startup overhead and data transmission delay, and provides an efficient general parallel solution for high-dimensional tensor computing.

\subsubsection{CUDA-based nonlinear interpolation rotation}

Mathematically, the image blurring function outputs the final pixel value of the image by summing the weighted pixel blocks around each pixel in the input image. This section employs CUDA-based nonlinear interpolation rotation method (see Figure \ref{fig:overall_design} GPU rotation diagram). Figure \ref{fig:cuda3} shows an example of image blur using 3 $\times$ 3 pixel blocks. CUDA accelerates computation by assigning image pixels to threads for parallel processing. In rotation operations, pixels in different areas (such as corners, edges, and interiors) may involve different boundary conditions or neighbourhood access patterns during computation, which may require targeted optimisation. In the figure below, a, b, and c correspond to three typical regions of the image. The analysis, combined with CUDA thread blocks (illustrated as 3 $\times$ 3 thread blocks) and dedicated kernel function design, is as follows:

\begin{figure*}[t]
\centering
 \includegraphics[width=1\textwidth]{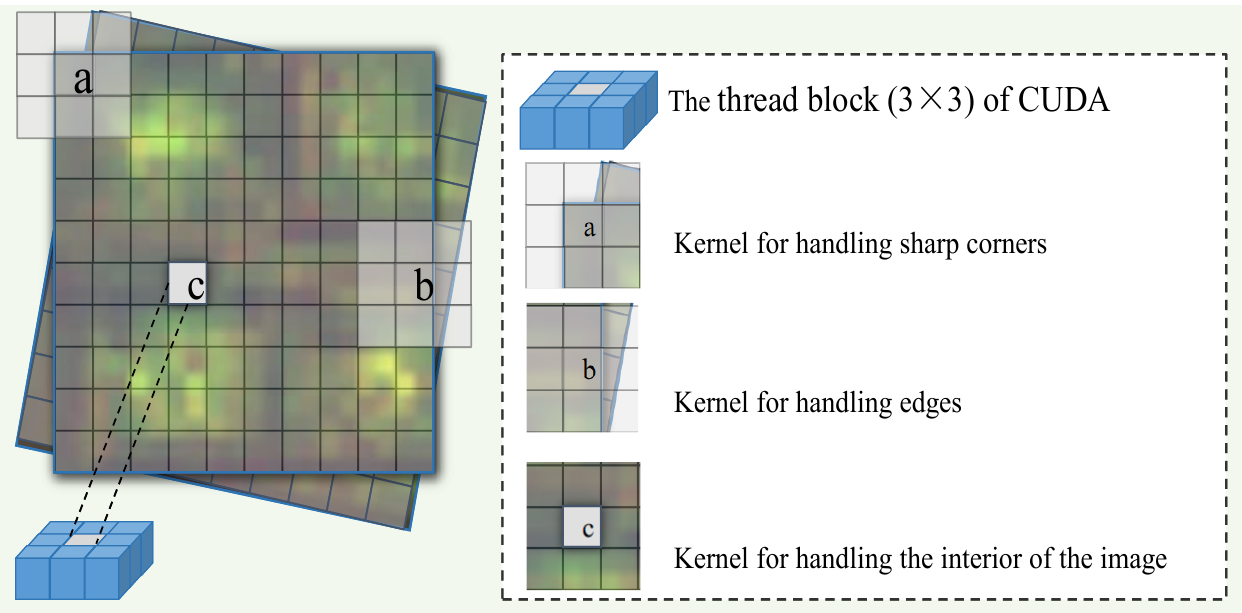}
   \caption{Processing rotaion conditions of pixels close to image edge.}
   \label{fig:cuda3}
\end{figure*}

\begin{itemize}
        \item[\textcolor{black}{$\bullet$}] Figure \ref{fig:cuda3} legend a: The "sharp corners" of the image. The four corner areas formed after image rotation (such as the top left corner, top right corner, etc.) are the "vertex" regions of the rotated image. Corner pixels may correspond to extreme edge positions in the original image before rotation, and coordinate calculations can easily exceed the boundaries of the original image, requiring special handling (such as boundary padding, cropping, or interpolation). The number of pixels in the corner areas is small, but neighbourhood information may be incomplete (for example, there may be no corresponding pixels in the original image), necessitating separate handling of boundary conditions. The neighbourhood dependency of corner area pixels is low, and simple nearest-neighbour interpolation or zero padding can be used to reduce the computational cost of bilinear interpolation.
        \item[\textcolor{black}{$\bullet$}] Figure \ref{fig:cuda3} legend b: "Edge area" of the image. The pixel area of the four edges (excluding corners) of the image after rotation, located between the corners and the interior. Edge pixels may correspond to boundary rows or columns in the original image, and some neighbouring pixels may extend beyond the original image's boundaries (such as the top edge pixels of the rotated image, where there may be no upper pixels in the original image). The boundary is directionally filled according to the direction of the edge (e.g., only filling upper pixels for the top edge, only filling right-side pixels for the right edge) to avoid redundant calculations. Edge pixels may require bilinear interpolation to ensure smoothness, but the interpolation direction can be simplified according to the edge direction (e.g., horizontal edges only need vertical interpolation).
        \item[\textcolor{black}{$\bullet$}] Figure \ref{fig:cuda3} legend c: "Inner region" of the image. After rotation, the middle region of the image, excluding corners and edges, contains the most pixels. The neighbourhood data of pixels in the inner region are complete, making it suitable for complex interpolation (such as bilinear or bicubic interpolation), and the memory access pattern is regular (row and column continuous), with a high data reuse rate. This is processed in parallel by a large number of 3 $\times$ 3 thread blocks, with each thread block corresponding one-to-one with an image block (for example, each 3 $\times$ 3 thread block processes a 3 $\times$ 3 pixel block in the original image, which is then mapped to the inner region of the target image after rotation). High-precision interpolation (such as bilinear or bicubic) can be applied to the inner pixels, utilising the complete neighbourhood data in shared memory for parallel computation, enhancing the image quality.
\end{itemize}

The following code addresses the handling of boundary overflow caused by CUDA thread grids that do not necessarily cover the entire image. This ensures that images of any size can be completely covered (without missing the right or bottom boundary). Portions exceeding the access boundary are explicitly filtered, preventing the CUDA kernel function from crashing or producing erroneous results due to out-of-bound read or write operations. This approach is versatile in its handling.

\begin{lstlisting}[numbers=left]
int col = blockIdx.x * blockDim.x + threadIdx.x;
int row = blockIdx.y * blockDim.y + threadIdx.y;
if(col < w && row < h){
    int pixVal = 0;
    int pixels = 0;
}
for(int blurRow =-BLUR_SIZE;blurRow<BLUR_SIZE+1;++blurRow){
    for(int blurCol=-Blur_SIZE;blurCol<BLUR_SIZE+1;++blurCol){
        int curRow = row + blurRow;
        int curCol = col + blurCol;
        if(curRow>=0 && curRow<h && curCol>=0 && curCol<w){
           pixVal += in[curRow * w + curCol];
           ++pixels;
        }}}
        out[row*w+col] = (unsigned char(pixVal/pixels);
}
\end{lstlisting} 

In the code above, the linearised indices curRow and curCol are used to access the value of the input pixel accessed in the current iteration. They accumulate pixel values into a running sum variable pixval. Line 12 records the addition of another pixel value to the running sum by incrementing the pixels variable. After processing all the pixels in the block, the average pixel value in the block is calculated by dividing the pixval value by the pixel value in line 15, and the result is written to the output pixel using the linearised indices of the row and column. Line 3 contains a conditional statement to prevent outbound access. As shown in Figure \ref{fig:cuda3}, when calculating output pixels near the edge of the image, the pixel block may exceed the valid range of the input image. In this case, the row and column values of the output pixel are 0 and 0, respectively. During the execution of the nested loops, it is observed that for five pixels outside the image, at least one value is less than 0. The conditions in the if statement, currow<0 and curcol<0, are responsible for catching these values. Therefore, only the values of 4 valid pixels are added to the running sum variable, and the value of the pixels is correctly incremented only 4 times to accurately compute the average in line 15. Most threads will find all pixels in their assigned 3 $\times$ 3 $\times$ 3 pixel block. They will accumulate all 9 pixels. However, for the four corner pixels, the responsible thread will accumulate only 4 pixels.

\subsection{Our key contribution}
The proposal of an "end-to-end" CUDA parallel deblurring strategy for wind-induced blur aims to address the problem of blur errors caused by wind disturbances in agricultural images. By integrating dynamic blur-robust convolution, the YOLOv11-DFRC model's real-time recognition in blurred scenarios is enhanced, providing the possibility for real-time detection on terminal devices, as detailed below:
\begin{itemize}
    \item[\textcolor{black}{$\bullet$}] "Boundary Safe Mean Filtering" and DFRC have been integrated into a single-core dual-channel pipeline within a CUDA kernel. Channel 1 performs lightweight mean filtering to quickly eliminate high-frequency wind disturbance streaks; channel two uses learnable weights $\alpha\in[0,1]$ to blend the filtering results with the original image per pixel, achieving a differentiable adjustment of deblurring intensity. The value of $\alpha$ is generated through online feedback of the detection confidence of YOLOv11-DFRC, forming a closed loop of "detection-deblurring".
    \item[\textcolor{black}{$\bullet$}] Real-time metrics: In terms of latency, for a 640 $\times$ 680 input, the delay in the DFRC unit is 6.4 ms. In terms of energy efficiency, compared to the traditional Gaussian pyramid deblurring scheme of the CPU, the FPS is improved by 5.8×, and energy consumption per frame is reduced by 62\%.
\end{itemize}

\section{Experiments and Analysis}
\subsection{Experimental environment and dataset}
These data sets are collected mainly in the Changji region of Xinjiang, which is influenced by continental climate and terrain, and seasonal changes also have a significant impact on wind. Therefore, the wind also has a significant impact on the collection and identification of wheat pests and diseases, which can easily cause ghosting and blurring. Tags are used to label over 3900 images. The experimental environment uses a 4-card Quadro RTX 6000, and the data set is divided into training, testing, and validation components in a ratio of 7: 2: 1. The input images are compressed to 640 $\times$ 640, and the training epochs = 500, batch = 64, workers = 10, optimiser = SGD, patience = 30.

In order to evaluate the effect of the DFRC unit on YOLOv11 in fuzzy scenes and non-fuzzy scenes, the following ablation experiments were performed to compare them, as shown in Table \ref{tab:Results of the ablation experiment}:

\begin{table*}[t]
  \centering 
  \captionsetup{justification=centering}
  \caption{Performance comparison analysis of different models.}
  \label{tab:Results of the ablation experiment}
  \begin{tabular}{ *{5}{c} }
    \toprule 
    \textbf{experiment} & \textbf{DFRC unit} & \textbf{Fuzzy data set} & \textbf{FPS} & \textbf{mAP50} \\
    \midrule    
    EfficientDet\cite{tan2020efficientdet,xia2024research} &  & $\times$ & 23 & 24.2\% \\
    EfficientDet\cite{11130331,10757874} &  & \checkmark & 22 & 33.8\% \\
    RetinaNet\cite{feng4681618enhancing} &  & $\times$ & 18 & 20.5\% \\
    RetinaNet\cite{nandibewoor2023computer} &  & \checkmark & 15 & 39.1\% \\
    SSD\cite{yan2022using} &  & $\times$ & 59 & 55.9\% \\
    SSD\cite{martins2025enhancing} &  & \checkmark & 50 & 77.2\% \\
    SPP-net\cite{kiran2025spatial,wu2025solidtrack} &  & $\times$ & 20 & 42.4\% \\
    SPP-net\cite{zhang2020insulator} &  & \checkmark & 24 & 63.1\% \\
    R-CNN\cite{sayed2021improved} &  & $\times$ & 11.4 & 57.1\% \\
    R-CNN\cite{10249515} &  & \checkmark & 15.5 & 66.4\% \\
    Fast RCNN\cite{wen2021video} &  & $\times$ & 20 & 51.3\% \\
    Fast RCNN\cite{xu2016real} &  & \checkmark & 23 & 70.0\% \\
    Faster R-CNN\cite{yang2018application} &  & $\times$ & 21 & 54.1\% \\
    Faster R-CNN\cite{olorunshola2023comparative} &  & \checkmark & 25 & 73.2\% \\
    YOLOv8\cite{zheng2021deblur}  &  & $\times$ & 37 & 68.7\% \\
    YOLOv8\cite{okur2024two}  &  & \textbf{\checkmark} & 41 & 43.3\% \\
    
    YOLOv11 (baseline) & $\times$ & $\times$ & 34 & 71.2\% \\
    \textbf{YOLOv11-DFRC} & \textbf{\checkmark} & \textbf{$\times$} & \textbf{45} & \textbf{79.4\%} \\
    \textbf{YOLOv11-DFRC} & \textbf{\checkmark} & \textbf{\checkmark} & \textbf{45} & \textbf{88.9\%} \\
    YOLOv11 (baseline) & $\times$ & Rainy weather & 45 & 58.4\% \\
    \textbf{YOLOv11-DFRC} & \textbf{\checkmark} & \textbf{Rainy weather} & \textbf{46} & \textbf{79.5\%} \\
    \bottomrule 
  \end{tabular}
\end{table*}

Table \ref{tab:Results of the ablation experiment} presents a comparative experiment for the proposed YOLOv11-DFRC framework in wheat pest and disease detection, focusing on evaluating the impact of DFRC and fuzzy data augmentation. Key findings are summarized as follows:
First,general Impact of Fuzzy Datasets on Model Performance. As shown in the table, all models exhibit varying degrees of improvement in mAP50 on the fuzzy dataset (marked \checkmark), but FPS generally decreases. For example: EfficientDet achieves mAP50 of 24.2\% on the clear dataset ($\times$) and improves to 33.8\% on the fuzzy dataset, but FPS drops from 23 to 22. SSD achieved mAP50 of 55.9\% on the clear dataset, surged to 77.2\% on the fuzzy dataset, but FPS decreased from 59 to 50. This indicates that fuzzy features contain more target detail information (e.g., blurred edges of pest/disease contours), but the computational complexity of feature extraction also increases, leading to reduced inference speed.Secend, differences in Blur Robustness Across Detection Frameworks Single-stage detectors: SSD and YOLOv11-DFRC demonstrate outstanding performance in blurred scenarios. SSD achieved a 21.3\% mAP50 improvement, while YOLOv11-DFRC reached an mAP50 of 88.9\% on the blurred dataset with FPS maintained at 45. This demonstrates the efficiency advantage of single-stage models in handling blurred features through their “end-to-end” detection approach. Two-stage detectors: The R-CNN series showed limited mAP50 improvement on blurred datasets (e.g., R-CNN increased from 57.1\% to 66.4\%, a 9.3\% gain), with negligible FPS gains (R-CNN rose from 11.4 to 15.5). This indicates that the two-stage “candidate region followed by classification/regression” workflow is less efficient at utilizing blurred features. Anchor-Free Models: EfficientDet and RetinaNet exhibit the lowest absolute mAP50 values, failing to exceed 40\% even on blurry datasets. This reflects anchor-free frameworks' inadequate localization capabilities for small and blurry targets, demonstrating poor adaptability in scenarios like agricultural pest detection where objects are “small and marginally blurred.”
Third, performance Advantages of YOLOv11-DFRC Using the YOLOv11 baseline model as a reference, we analyze the improvements achieved by YOLOv11-DFRC:
On clear datasets: YOLOv11-DFRC increases mAP50 from 71.2\% to 79.4\% and FPS from 34 to 45, demonstrating that the DFRC unit enhances feature expression and improves inference efficiency even in clear scenarios. Blurred Datasets: mAP50 surged to 88.9\%, a 30.5\% improvement over the baseline, demonstrating the DFRC unit's strong robustness against motion blur. Rainy blur scenarios: YOLOv11-DFRC achieves mAP50 of 79.5\%, significantly surpassing the baseline's 58.4\%, while maintaining 46 FPS, demonstrating its generalization capability across diverse blur environments. 
Thus,in the dynamic blur detection scenario for wheat pests and diseases in agriculture, YOLOv11-DFRC outperformed other mainstream detection models in detection accuracy (mAP50), real-time performance (FPS) and environmental generalization, leveraging the feature enhancement capabilities of the DFRC unit. In contrast, two-stage models and anchor-free models showed weaker performance in blurred scenarios and are better suited for object detection tasks in clear environments.

\begin{table*}[t]
  \centering 
  \captionsetup{justification=centering}
  \caption{detection results comparison of different models.}
  \label{tab:detection_comparison}
  \small 
  \setlength{\tabcolsep}{4pt} 
  \begin{tabular}{lccc|lccc}
    \toprule
    \multicolumn{4}{c|}{\textbf{YOLOv11 (Baseline)}} & \multicolumn{4}{c}{\textbf{YOLOv11-DFRC}} \\
    \midrule
    Class & P & R & mAP50 & Class & P & R & mAP50 \\
    \midrule
    all & 0.762 & 0.816 & 0.555 & all & 0.807 & 0.859 & 0.862 \\
    Bacterial\_Leaf\_Blight & 0.873 & 0.898 & 0.643 & Bacterial\_Leaf\_Blight & 0.701 & 0.721 & 0.794 \\
    Brown\_Spot & 0.692 & 0.773 & 0.429 & Brown\_Spot & 0.719 & 0.738 & 0.783 \\
    HealthyLeaf & 0.560 & 0.617 & 0.440 & HealthyLeaf & 0.738 & 0.542 & 0.724 \\
    Leaf\_Blast & 0.760 & 0.765 & 0.488 & Leaf\_Blast & 0.693 & 0.783 & 0.764 \\
    Leaf\_Scald & 0.792 & 0.871 & 0.618 & Leaf\_Scald & 0.725 & 0.789 & 0.790 \\
    Narrow\_Brown\_Leaf\_Spot & 0.730 & 0.784 & 0.535 & Narrow\_Brown\_Leaf\_Spot & 0.714 & 0.759 & 0.765 \\
    Neck\_Blast & 0.850 & 0.921 & 0.664 & Neck\_Blast & 0.811 & 0.875 & 0.741 \\
    Rice\_Hispa & 0.840 & 0.895 & 0.627 & Rice\_Hispa & 0.853 & 0.868 & 0.737 \\
    \bottomrule
  \end{tabular}
\end{table*}

Table \ref{tab:detection_comparison} shows that the YOLOv11 baseline model demonstrates high detection accuracy in disease categories such as Bacterial\_Leaf\_Blight, Neck\_Blast, and Rice\_Hispa, exhibiting good recognition capability of typical symptom morphology. However, its detection performance for the HealthyLeaf and Brown\_Spot categories is noticeably lower, revealing the model's inadequacies in distinguishing healthy leaves from mild lesions, low-contrast texture changes, and subtle lesion areas, reflecting the current feature extraction mechanism's limitations in fine-grained discrimination. The experimental results of the YOLOv11-DFRC model in the table show that the overall detection performance of the improved model tends to be more balanced, with an mAP50 of 0.462, precision P of 0.707, and recall rate R of 0.759, demonstrating better detection stability and generalisability. In particular, it performs outstandingly in the Neck\_Blast and Rice\_Hispa categories, with mAP50 reaching 0.541 and 0.537, respectively, and precision exceeding 0.8, while recall surpasses 0.85, indicating that structural optimisation and enhanced attention mechanisms have effectively improved the localisation and classification accuracy of key diseases. However, the model still experiences omissions and misjudgments when dealing with blurred edges, similar colours, or small-scale lesions (such as early Brown\_Spot), suggesting that future work needs to further strengthen multiscale feature fusion capabilities, enhance contextual awareness, and introduce more refined boundary modelling strategies to improve the discriminative performance for categories with minimal visual differences. 

\begin{table*}[t]
  \centering
  \captionsetup{justification=centering}
  \caption{Comparison of experimental results based on the improved YOLOv11-DFRC module.}
  \label{tab:Comparison of experimental results based on the improved YOLOv11-DFRC module}
  \begin{tabular}{cccccc}
    \toprule 
    Model & Fuzzy data set & mAP50 & FLOPs & Training time\\
    \midrule
    YOLOv11 (baseline) & $\times$ & 71.2\% & 34 & 1.2 hour\\
    YOLOv11 (baseline) & \checkmark & 62.8\% & 36 & 1.2 hour\\
    YOLOv11-DFRC & $\times$ & 79.4\% & 45 & 1.6 hour\\
    YOLOv11-DFRC & \checkmark & 88.9\% & 45 & 1.6 hour\\
    YOLOv11-DFRC $+$ CUDA & $\times$ & 79.4\% & 38 & 1.3 hour\\
    YOLOv11-DFRC $+$ CUDA & \checkmark & 88.9\% & 38 & 1.3 hour\\
    YOLOv11 (baseline) $+$ DRS & $\times$ & 84.8\% & 31 & 1.6 hour\\
    YOLOv11 (baseline) $+$ DRS & \checkmark & 86.1\% & 31 & 1.6 hour\\
    YOLOv11 (baseline) $+$ DRS $+$ CUDA & $\times$ & 86.1\% & 32 & 1.3 hour\\
    DFRC-CPU-unit & $\times$ & $-$ & $-$ & 688.45 ms\\
    DFRC-GPU-unit & $\times$ & $-$ & $-$ & 17.94 ms\\
    \bottomrule 
  \end{tabular}
\end{table*}

As shown in Table \ref{tab:Comparison of experimental results based on the improved YOLOv11-DFRC module}, the comparative analysis of experimental results based on the improved YOLOv11-DFRC module is as follows:
First, regarding the core performance enhancement of the DFRC module, comparing the YOLOv11 baseline model with YOLOv11-DFRC. On the clear dataset (\checkmark), mAP50 improved from 71.2\% to 79.4\%, while FPS increased from 34 to 45. This demonstrates that the DFRC module can enhance feature representation and optimize inference efficiency even in clear scenes. For the blurry dataset (\checkmark), mAP50 performance surged from 62.8\% to 88.9\%, representing a 26.1\% improvement. This demonstrates the DFRC module's strong adaptability to dynamic blur features, serving as the key to the model's accuracy breakthrough in blurry scenarios. Second, regarding CUDA acceleration's impact on efficiency and runtime, we analyzed the inference speed (FPS) of YOLOv11-DFRC versus YOLOv11-DFRC + CUDA. Adding CUDA reduced FPS from 45 to 38, yet it remained higher than the YOLOv11 baseline of 34/36. Regarding training time, the duration shortened from 1.6 hours to 1.3 hours. Combined with the inference latency comparison between DFRC-CPU-unit and DFRC-GPU-unit (688.45 ms vs. 17.94 ms), this demonstrates that CUDA parallel acceleration significantly reduces the computational overhead of the DFRC unit, addressing real-time performance and training efficiency bottlenecks on edge devices. Third, regarding feature fusion gains in the DRS module, YOLOv11 (baseline) + DRS achieved an mAP50 improvement from 71.2\% to 84.8\% on the clear dataset, while FPS decreased from 34 to 31. This indicates that DRS's “dynamic weight fusion” strategy enhances feature expression but incurs computational overhead. On the blurry dataset, mAP50 improved from 62.8\% to 86.1\%. While slightly below YOLOv11-DFRC's 88.9\%, this demonstrates adaptability to blurry features. When combined with CUDA, (YOLOv11 (baseline) + DRS + CUDA), FPS recovered to 32, and training time was reduced to 1.3 hours, demonstrating that the synergy between DRS and CUDA optimizes efficiency.
Thus, YOLOv11-DFRC achieves multidimensional optimization for dynamic blur detection of agricultural pests and diseases through: DFRC module enhancing fuzzy feature extraction - CUDA acceleration optimizing efficiency and runtime DRS module assisting feature fusion Specifically, DFRC serves as the core for accuracy enhancement, CUDA is pivotal for efficiency optimization, and DRS provides dynamic adaptability for feature fusion. The synergistic integration of these three components enables the model to achieve industry-leading detection performance in blurred scenarios.

Next, we analyse the effect of DFRC through visualisation, with the visual results shown in Figure \ref{tab:Visual_heatmap}. It is evident that the model's attention range on blurred images is greater than the diffusion range of the attention area in nonblurred images, which proves that constructing DFRC at different feature levels is beneficial for global jitter learning. Furthermore, the FPN fusion of different levels of features can compensate for the neglected features.

\begin{table*}[t]
  \centering 
  \captionsetup{justification=centering, font=small, labelfont=bf}
  \caption{\textbf{Visual Comparison of Heatmaps under Various Environmental Conditions.}}
  \label{tab:Visual_heatmap}
  \renewcommand{\arraystretch}{1.4}
  \setlength{\tabcolsep}{2pt}
 
  \begin{tabular}{
    >{\centering}m{4.0cm}     
    >{\centering\arraybackslash}m{2cm}  
    >{\centering\arraybackslash}m{2cm}  
    >{\centering\arraybackslash}m{2cm}  
  }
    \toprule 
    \textbf{Environmental Condition} & 
    \textbf{Input Image} & 
    \textbf{YOLOv11} & 
    \textbf{YOLOv11-DFRC} \\
    \midrule
 
    Original & 
    \includegraphics[width=1.5cm, height=1.5cm, valign=c]{{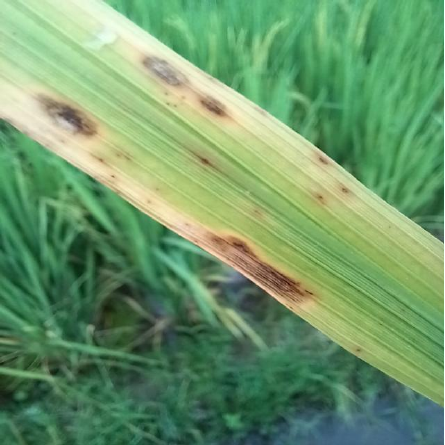}}  &
    \includegraphics[width=1.5cm, height=1.5cm, valign=c]{{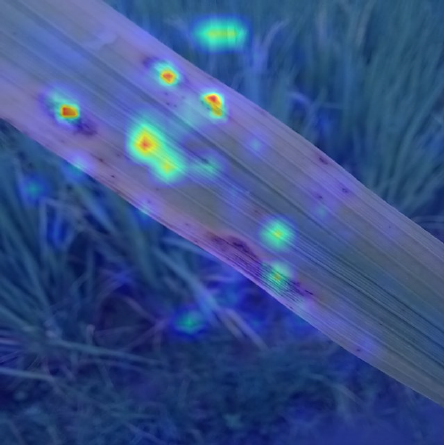}}  &
    \includegraphics[width=1.5cm, height=1.5cm, valign=c]{{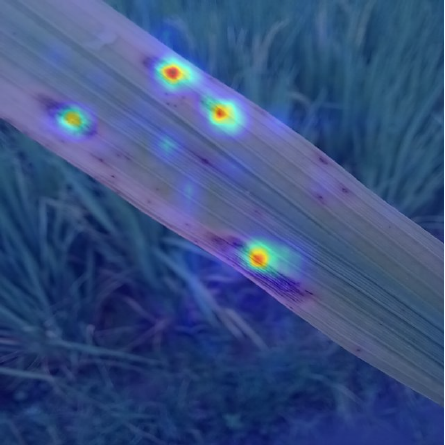}}  \\ 
 
    \addlinespace
 
    Blur (5px) & 
    \includegraphics[width=1.5cm, height=1.5cm, valign=c]{{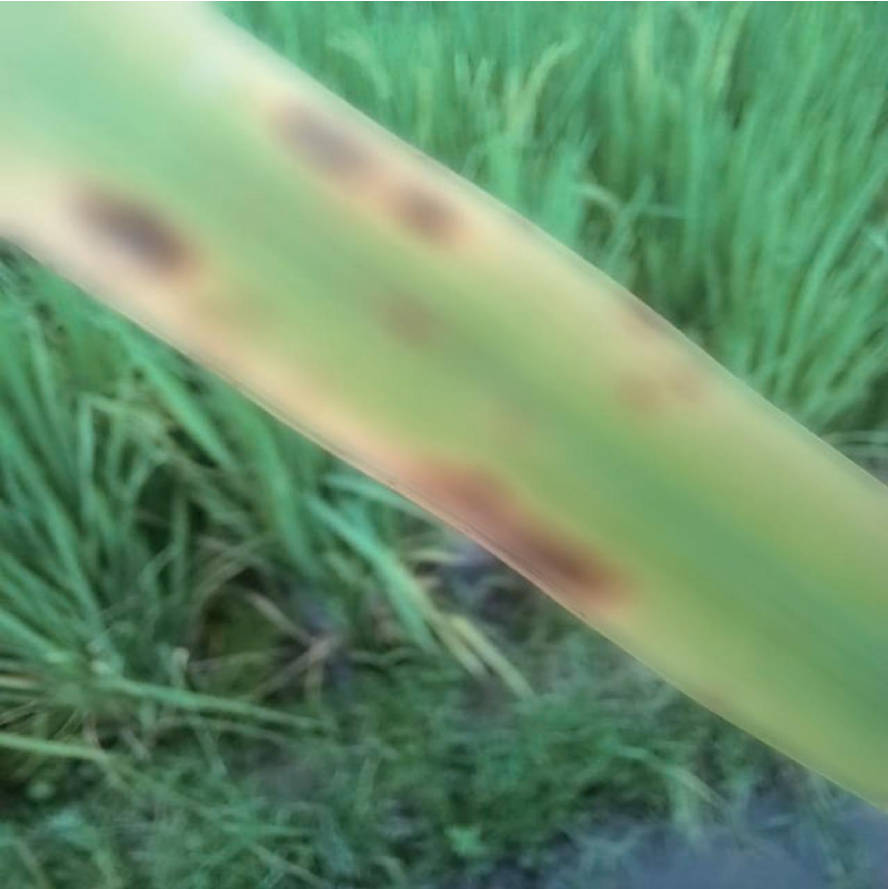}}  &
    \includegraphics[width=1.5cm, height=1.5cm, valign=c]{{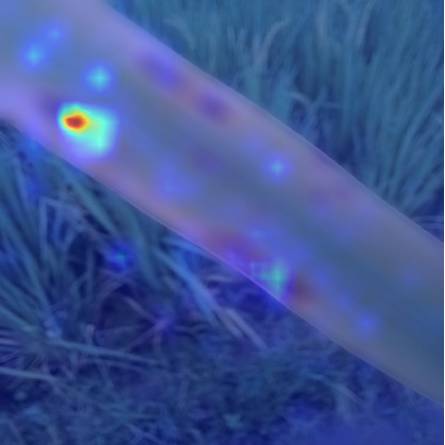}}  &
    \includegraphics[width=1.5cm, height=1.5cm, valign=c]{{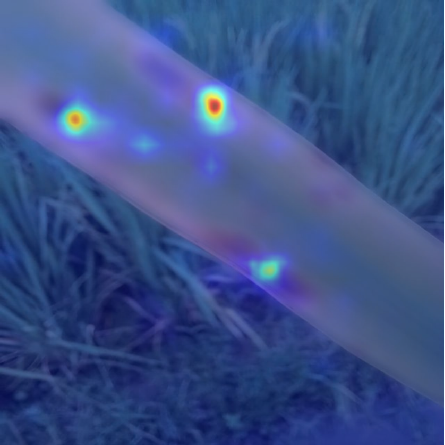}}  \\
 
    Blur (25px) & 
    \includegraphics[width=1.5cm, height=1.5cm, valign=c]{{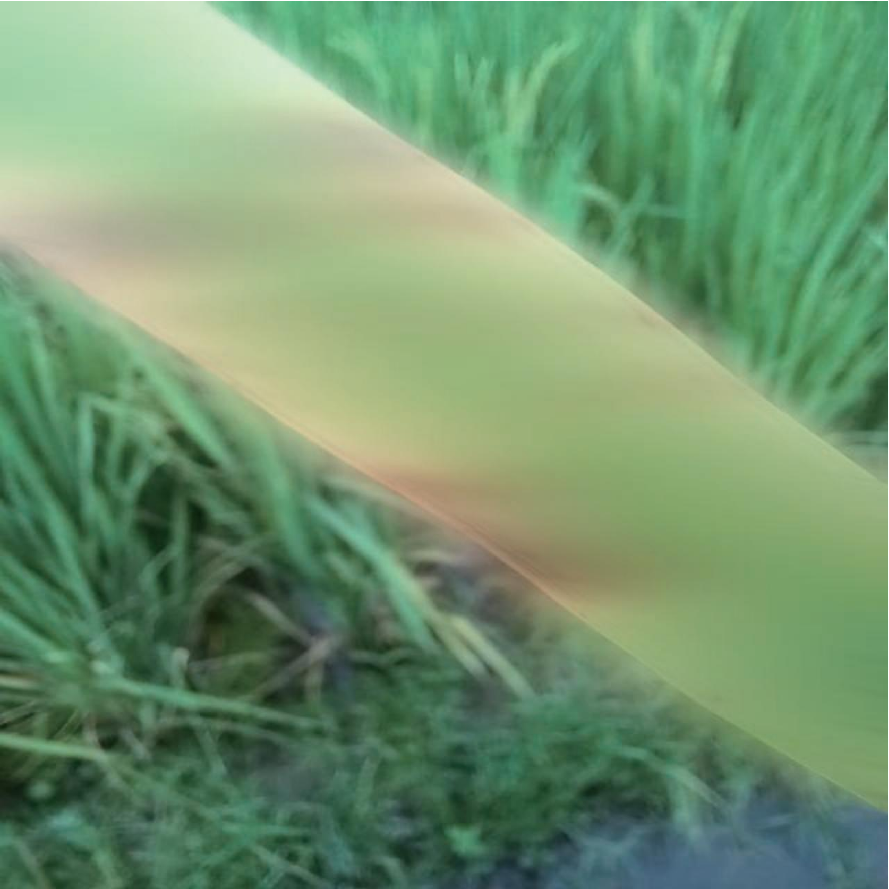}}  &
    \includegraphics[width=1.5cm, height=1.5cm, valign=c]{{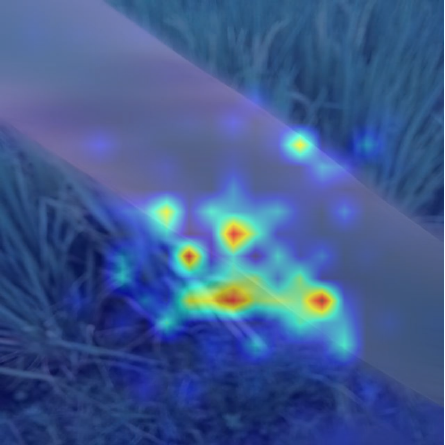}}  &
    \includegraphics[width=1.5cm, height=1.5cm, valign=c]{{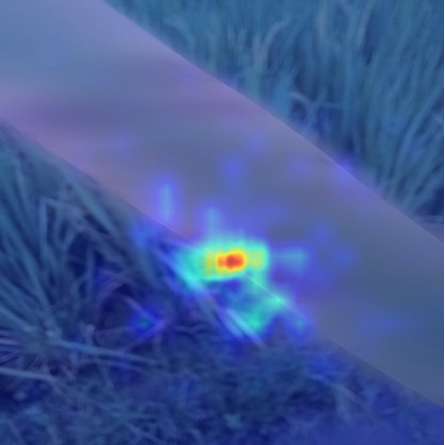}}  \\ 
 
    \addlinespace
 
    Cloudy (Original) & 
    \includegraphics[width=1.5cm, height=1.5cm, valign=c]{{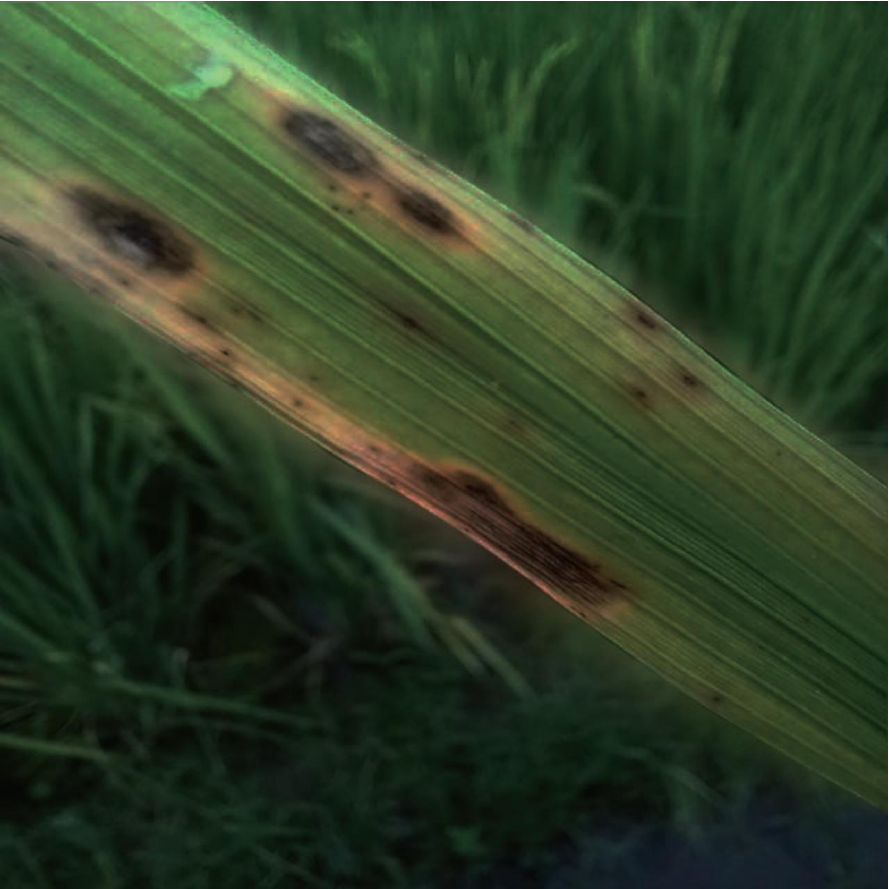}}  &
    \includegraphics[width=1.5cm, height=1.5cm, valign=c]{{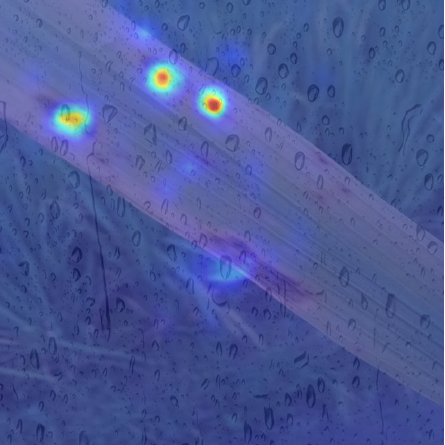}}  &
    \includegraphics[width=1.5cm, height=1.5cm, valign=c]{{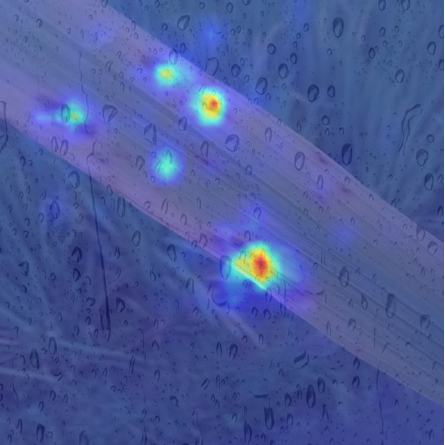}}  \\
 
    Cloudy + Blur (5px) & 
    \includegraphics[width=1.5cm, height=1.5cm, valign=c]{{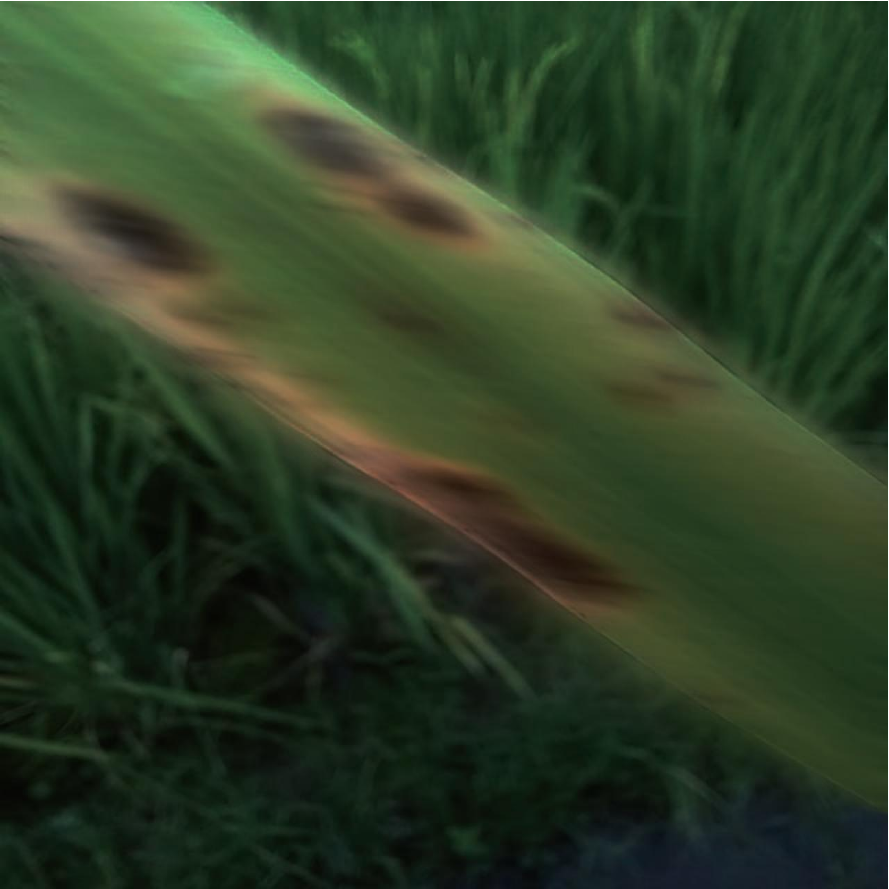}}  &
    \includegraphics[width=1.5cm, height=1.5cm, valign=c]{{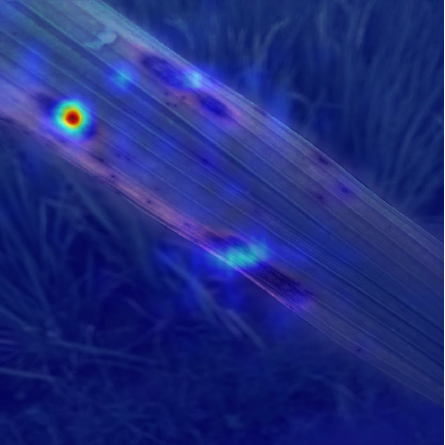}}  &
    \includegraphics[width=1.5cm, height=1.5cm, valign=c]{{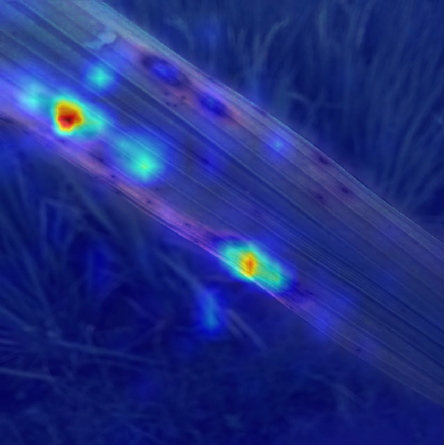}}  \\
 
    Cloudy + Blur (25px) & 
    \includegraphics[width=1.5cm, height=1.5cm, valign=c]{{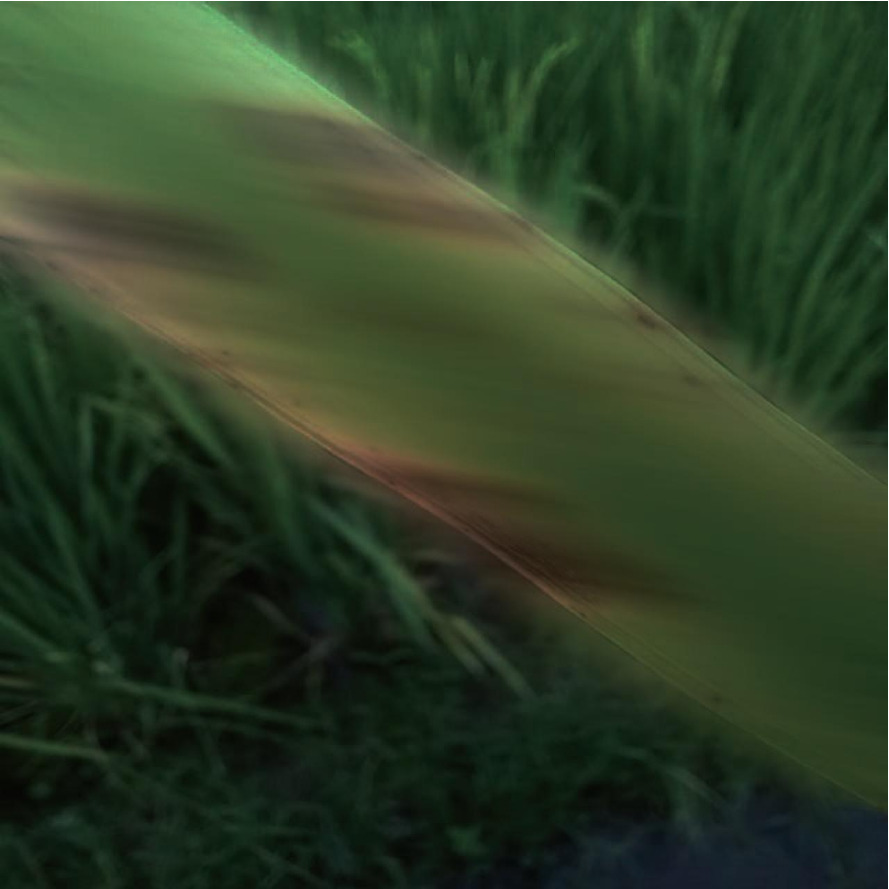}}  &
    \includegraphics[width=1.5cm, height=1.5cm, valign=c]{{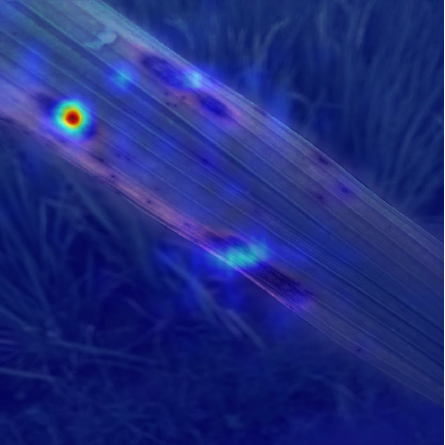}}  &
    \includegraphics[width=1.5cm, height=1.5cm, valign=c]{{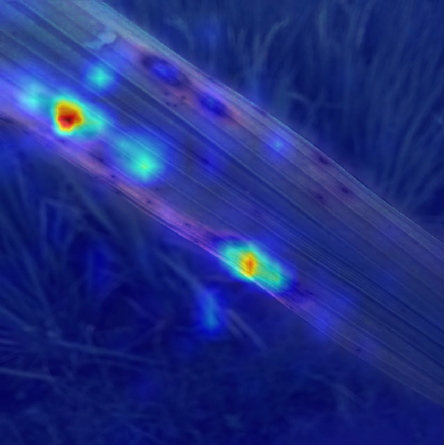}}  \\ 
 
    Raindrop Obscuration & 
    \includegraphics[width=1.5cm, height=1.5cm, valign=c]{{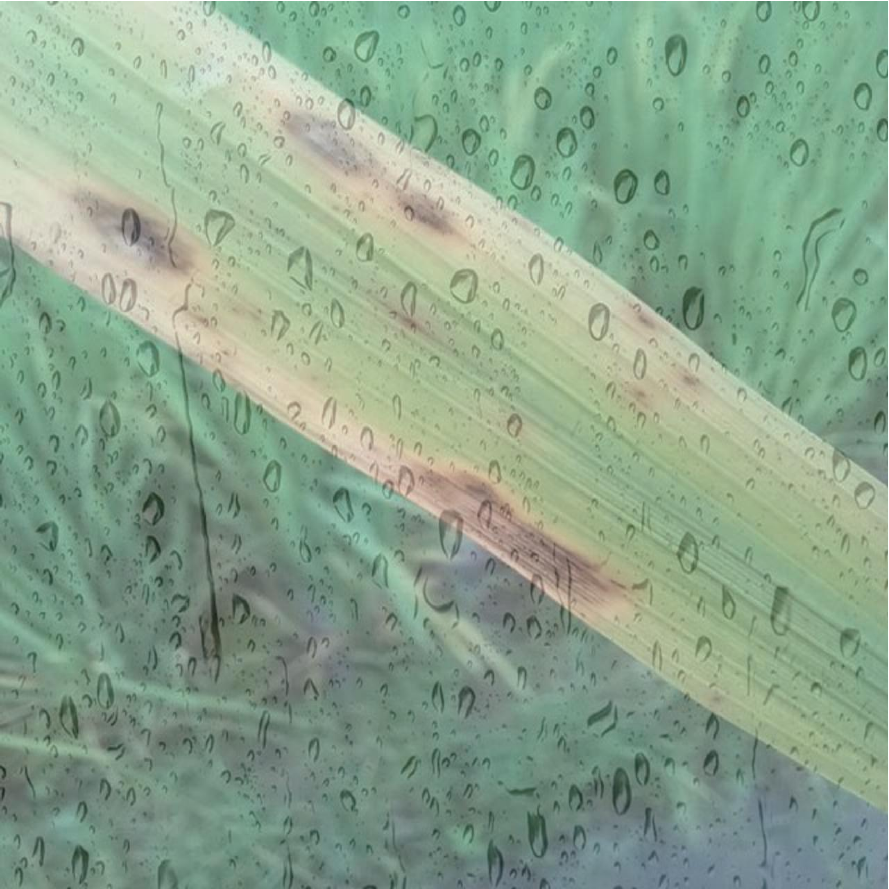}}  &
    \includegraphics[width=1.5cm, height=1.5cm, valign=c]{{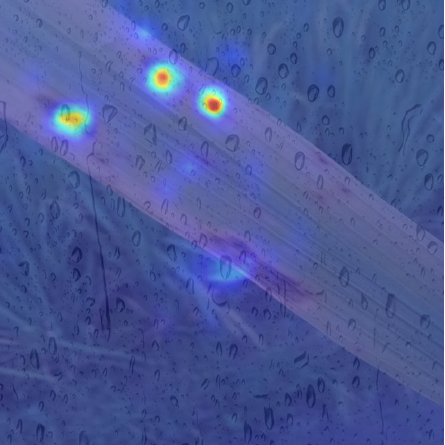}}  &
    \includegraphics[width=1.5cm, height=1.5cm, valign=c]{{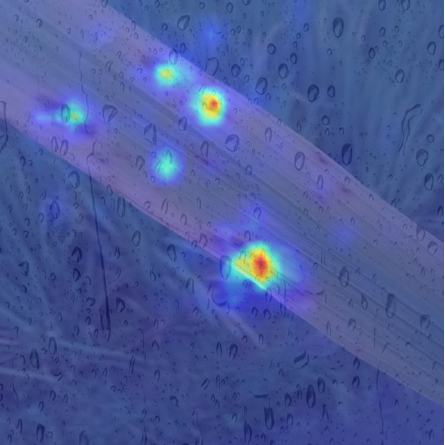}}  \\
 
    \bottomrule
  \end{tabular}
\end{table*}

By visualising thermal imaging, it is possible to intuitively understand the focus level of the YOLOv11-DFRC model in each area, indicating that the YOLOv11-DFRC model has a wider perception range for blurred targets. It shows excellent robustness and perceptual breadth in target detection tasks in complex agricultural environments. The visualised thermal maps clearly reveal that when dealing with blurred inputs, the model's attention is not only focused on the main body of the target but also effectively extends to its contextual areas, showing that DFRC enhances the model's spatial inference capability for blurred targets. Especially under real interferences such as motion blur and lens occlusion (such as raindrop coverage) , the model can still maintain high positioning sensitivity, while rainy and sunny weather has little impact on blurred recognition, verifying its adaptability under dynamic nonideal imaging conditions. The paired training strategy further strengthens the alignment of features between clear and blurred images, allowing the network to learn more invariant representations. Two methods of generating blurred data, global jitter blur and annotation box-based transparent fusion, systematically simulate dual interferences of equipment vibration and crop swinging, enhancing the model's generalisation performance in the ever-changing field environment. The role of the FPN structure in multiscale feature fusion is particularly critical as it effectively integrates low-level details and high-level semantic information, compensating for the loss of local features caused by blur, thereby achieving more complete target perception. This framework provides a feasible technical path for the stable operation of agricultural vision systems in situations of adverse weather or high-speed movement.

\begin{table*}[t]
  \centering
  \captionsetup{justification=centering}
  \caption{Two types of jitter data production.}
  \label{tab:Two types of jitter data production}
  \begin{tabular}{cc}
    \toprule
    Types of jitter blur & Description \\
    \midrule
    Uniform jitter blur & Generated based on simulation with blur kernels \\
    Bounding box jitter & Generated by perturbing bounding boxes \\
    \bottomrule
  \end{tabular}
\end{table*}

During the training process, paired training was adopted. In terms of dataset creation, two methods were used: 1. Uniform jitter blur was applied to blur the entire image to accommodate the jittering scenes of the shooting equipment. 2. Each target box labeled by labelimg was integrated through transparency to simulate motion blur caused by wind disturbances. The positions of the background area are preserved as much as possible. As shown in Table \ref{tab:Two types of jitter data production}, global controllable blur is used to simulate equipment vibration, enhancing the model's tolerance to noise from the shooting hardware. Local dynamic blur overlay preserves the background, enhancing the ability to extract features of natural movements such as wind disturbances, suitable for agricultural monitoring (crop wind disturbance) scenarios.

\begin{table*}[t]
  \centering
  \captionsetup{justification=centering}
  \caption{Two Types of Jitter Datasets with Visual Examples.}
  \label{tab:jitter_datasets}
  \begin{tabular}{@{}m{4cm}c>{\centering\arraybackslash}m{4cm}@{}}
    \toprule 
    \textbf{Blur Type} & \textbf{Fuzzy process} & \textbf{Description} \\
    \midrule 
    Type (1): \\
    Uniform jitter blur \\
    \adjustbox{valign=m}{\includegraphics[width=2cm]{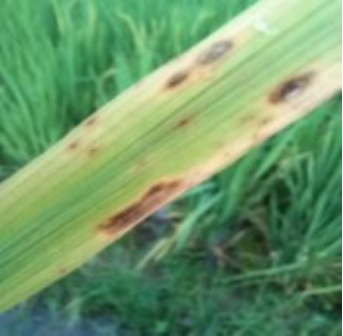}}  \quad & 
    \adjustbox{valign=m}
   {\includegraphics[width=2cm]{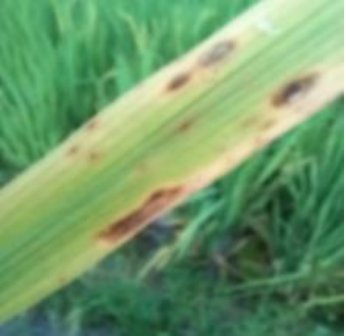}}  \quad &
    \parbox{4cm}{\begin{itemize}
      \item Uniform shaking blur is used to blur the entire image to adapt to the shaking scene of the shooting equipment.
      
    \end{itemize}} \\
    
    \addlinespace[0.5em]
    Type (2): \\
    Non-uniform jitter distributions \\
    \adjincludegraphics[width=1.8cm, valign=m]{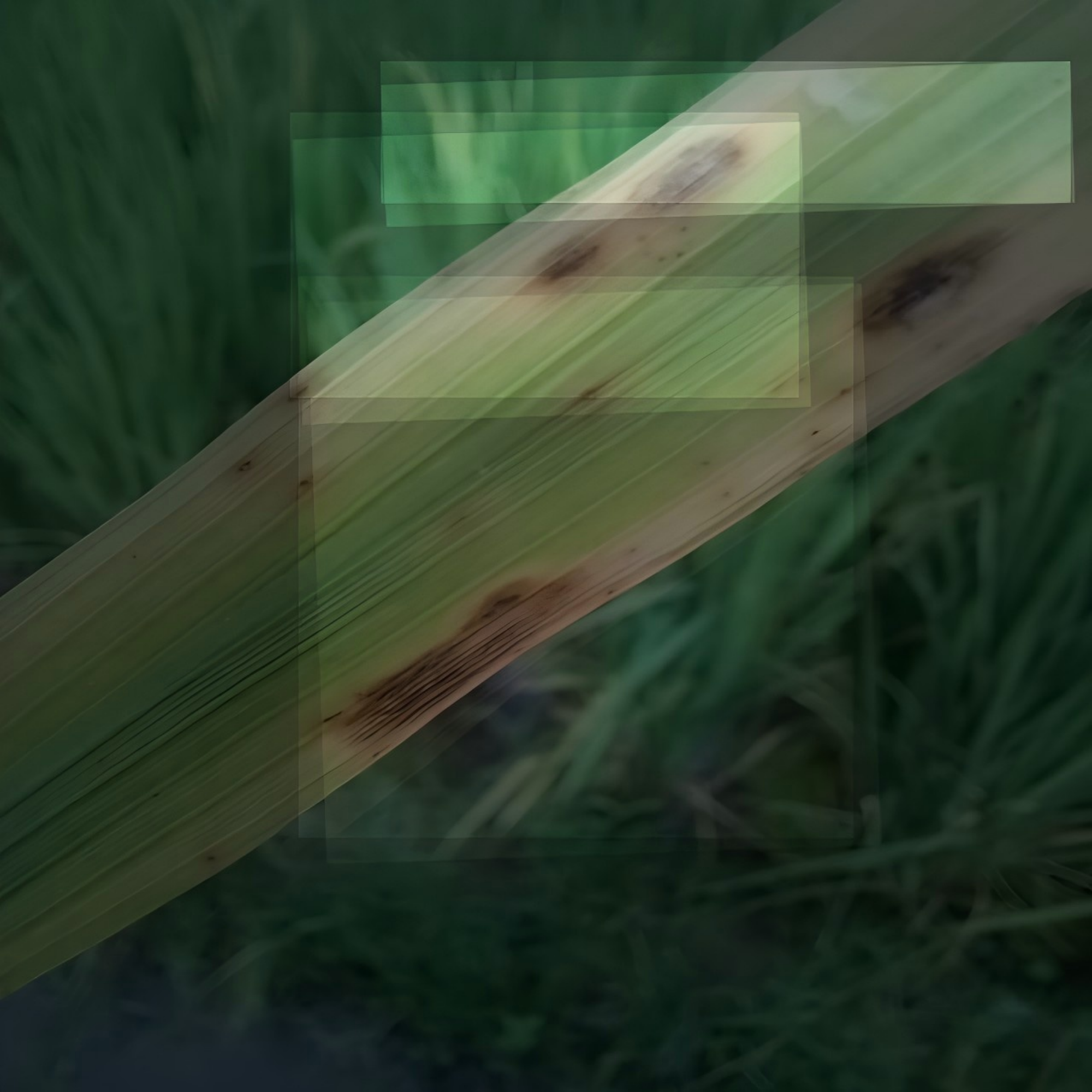} & 
    \begin{tabular}{@{}c@{}}
      
      \adjincludegraphics[width=1.8cm, valign=m]{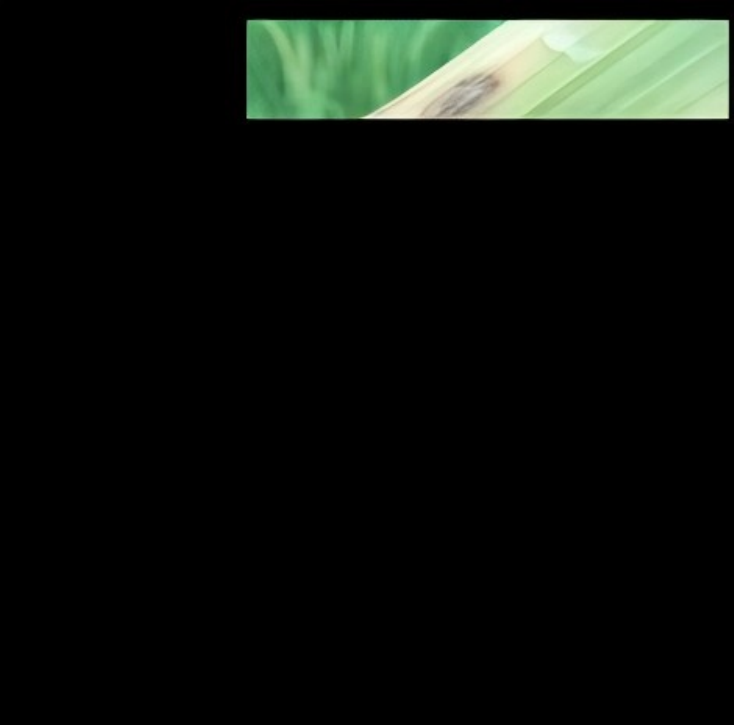}  \quad
      \adjincludegraphics[width=1.8cm, valign=m]{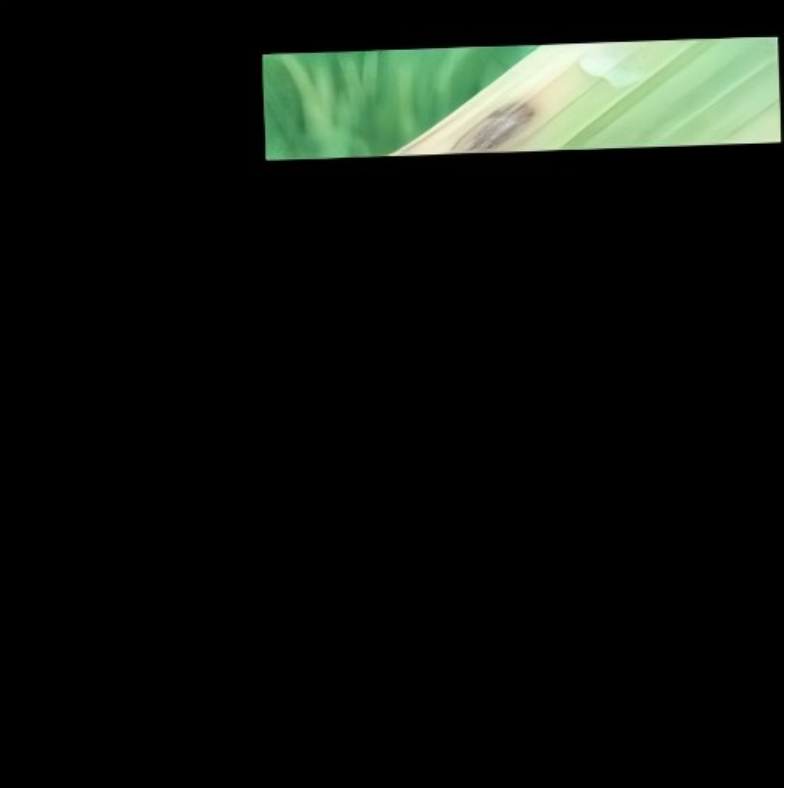}  \\
      \adjincludegraphics[width=1.8cm, valign=m]{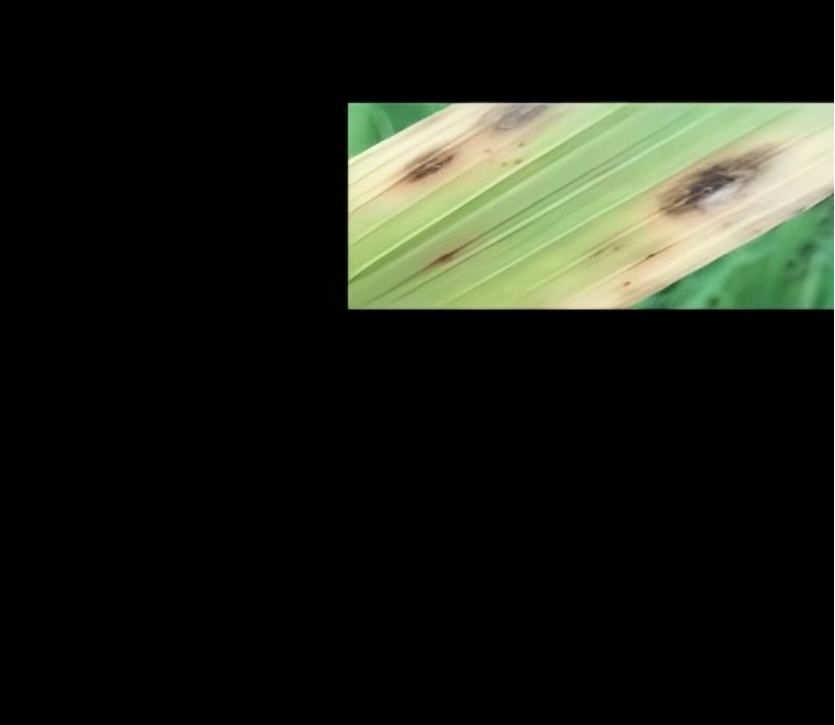}  \quad
      \adjincludegraphics[width=1.8cm, valign=m]{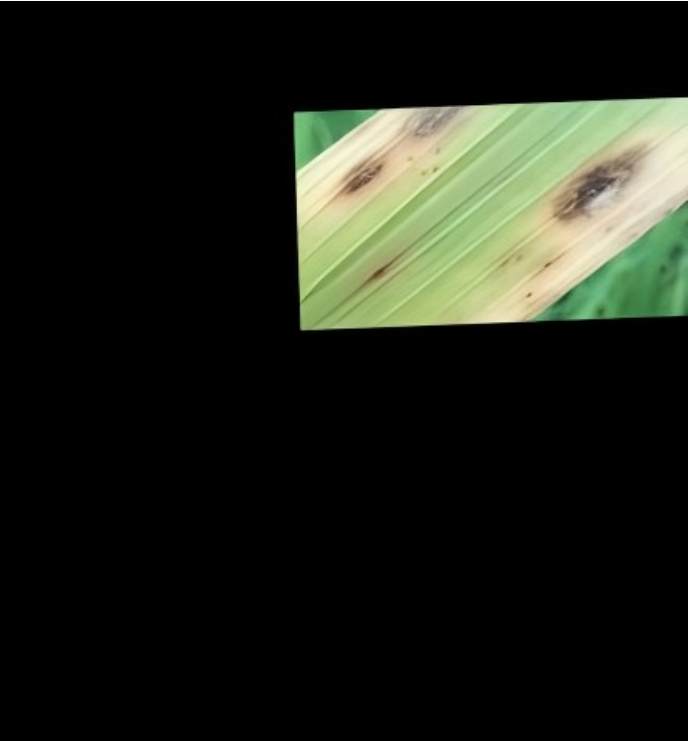}  \\
      \adjincludegraphics[width=1.8cm, valign=m]{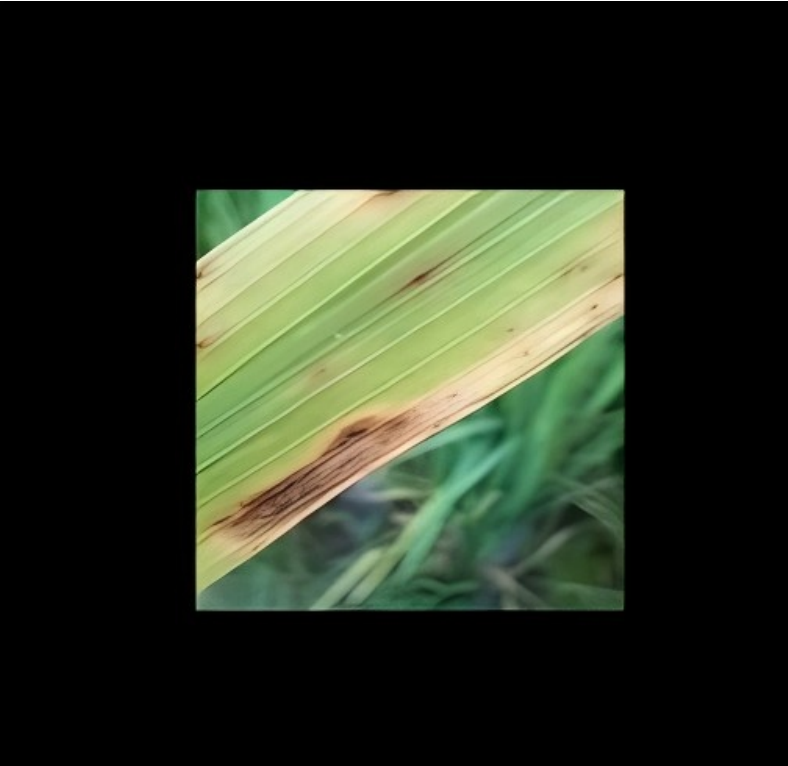}  \quad
      \adjincludegraphics[width=1.8cm, valign=m]{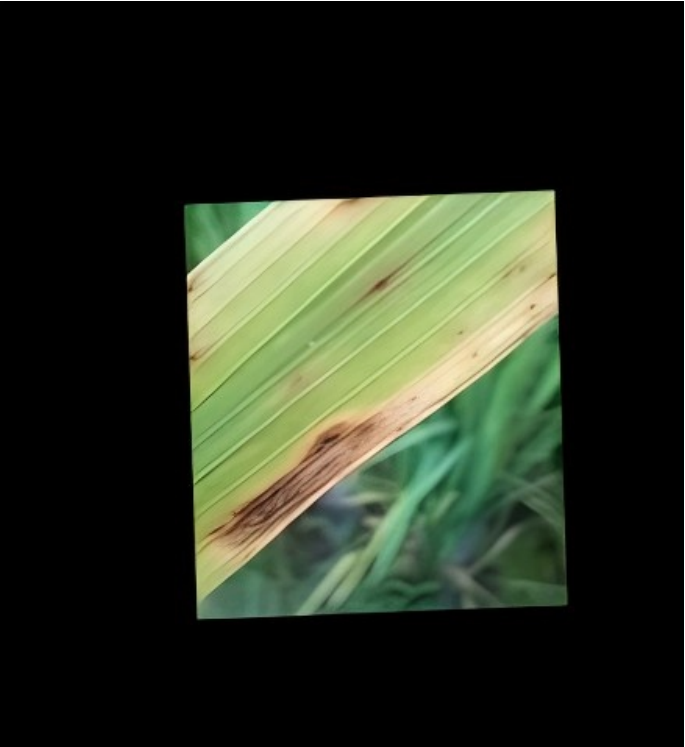}  \\
      
    \end{tabular} &
    \begin{itemize}
      \item  By integrating the bounding boxes through transparency, the model can pay more attention to the target jitter in non-background areas. 
      
    \end{itemize} \\
    \bottomrule
  \end{tabular}
\end{table*}

As shown in Table \ref{tab:jitter_datasets}, the dataset construction method adopts jitter blur and jitter blur within the bounding box datasets for paired training, which can simulate scenarios of displacement recognition caused by wind-induced disturbances to the camera or target, thereby enhancing the model's ability to recognise wind interference. By constructing the above two types of jitter blur datasets and conducting paired training, the model's adaptability and recognition robustness in wind interference scenarios can be significantly improved. The specific advantages are as follows:

\begin{itemize}
    \item[\textcolor{black}{$\bullet$}] Global uniform jitter blur (as shown in the first set of example images in the table) accurately simulates the blur effect caused by the overall shaking of the camera due to wind (such as drones or fixed monitoring equipment being displaced by wind) by applying a uniform jitter blur to the entire image. This results in all areas of the image (including both the background and the target) experiencing the blur effect simultaneously. This ensures that the model can adapt to the global blur interference caused by the device's own jitter.
    \item[\textcolor{black}{$\bullet$}] The non-uniform jitter blur within the bounding box (as shown in the second set of example images in the table) focuses on the motion of the target itself: by applying local jitter blur within the bounding box (target area), such as non-uniform blur in the leaf area, combined with transparency adjustment and sub-regional display (multi-subgraph comparison on the right side), it simulates the scene of the target (such as plant leaves or dynamic objects) swaying independently relative to the background under wind disturbance. This more closely aligns with the physical process of "wind-induced local displacement of the target," avoiding the interference of background blur on the target features.
\end{itemize}

By combining bounding boxes with a jittering blur process, the model can clearly distinguish between background and target areas during training, focusing attention on the non-background areas within the bounding box (such as diseased parts of leaves). This "target-oriented" blur strategy avoids feature dilution caused by global blur, allowing the model to prioritize learning subtle features of target jitter (such as displacement direction and magnitude) rather than background noise. Paired training (contrasting globally blurred and locally blurred samples) further enhances the model's ability to distinguish between "device jitter" and "target jitter," improving adaptability to complex wind field scenarios.

\section{Conclusion}
This study proposes a Dynamic Robust (DRFC) unit. This method transforms ghosting noise caused by camera and target jitter due to wind interference into a learnable feature. Through DRFC and a dynamic feature transparency mechanism, it achieves adaptive weighting and active perception fusion of jitter features, significantly reducing the workload of manual data cleaning in the early stages. A CUDA-based nonlinear interpolation rotation is designed, shifting fuzzy offset feature computation from CPU to GPU, and achieving a 400-fold efficiency improvement. Including DFRC into the model improved the accuracy of jitter perception by approximately 10\%. Experiments combining wind-induced jitter and raindrop-occluded lens observations demonstrate that DFRC units effectively resist jitter and raindrop interference, providing a plug-and-play blur processing module for dynamic agricultural vision tasks. Future research will explore feasible approaches for deploying anti-blur recognition on edge devices through multimodal fusion and distributed computing.\\

{\setlength{\parindent}{0pt} 

\textbf{Funding source:} This study was funded by the school-level project of Changji College, “Research on Introducing EfficientNet-B0 Lightweight YOLOv8 to Identify Wheat Pests and Diseases in Xinjiang” (Project No. KY2024017), and the Department of Education of the Xinjiang Uygur Autonomous Region, “Research on Talent Cultivation Mode of Software Engineering Specialties in the Applied Undergraduate Colleges and Universities under the Background of the New Engineering Science” (Project No. XJGXJGPTB-2024079) were jointly funded.\\

\textbf{Informed Consent Statement:} Informed consent was obtained from all subjects involved in the study.\\


\textbf{Acknowledgments:} The authors thank Changji College and and the Department of Education of the Xinjiang Uygur Autonomous Region for their continuous support in completing this work.\\

\textbf{Conflicts of Interest:} The authors declare no conflicts of interest.
}


\begin{thebibliography}{999}
\bibitem[\protect\citeauthoryear{Rai et al.}{{2024}}]{r1}
Rai, N., Zhang, Y., Villamil, M., Howatt, K., Ostlie, M., \& Sun, X. (2024). Agricultural weed identification in images and videos by integrating optimized deep learning architecture on an edge computing technology. \textit{Computers and Electronics in Agriculture}, \textit{216}, 108442. https://doi.org/10.1016/j.compag.2023.108442.

\bibitem[\protect\citeauthoryear{Huihui et al.}{{2023}}]{r2}
Huihui, Y., Daoliang, L., \& Yingyi, C. (2023). A state-of-the-art review of image motion deblurring techniques in precision agriculture. \textit{Heliyon}, \textit{9}(6), e17332. https://doi.org/10.1016/j.heliyon.2023.e17332.

\bibitem[Sun et al.(2025)]{Sun2025WaveletintegratedDN}
Sun, R., Li, X., Zhang, L., Su, Y., Di, J., \& Liu, G. (2025). Wavelet-integrated deep neural network for deblurring and segmentation of crack images. \textit{Mechanical Systems and Signal Processing}. https://api.semanticscholar.org/CorpusID:274927232.


\bibitem[Xiang et al.(2024)]{Xiang2024ApplicationOD}
Xiang, Y., Zhou, H., Li, C., Sun, F., Li, Z., \& Xie, Y. (2024). Application of Deep Learning in Blind Motion Deblurring: Current Status and Future Prospects. \textit{ArXiv}, abs/2401.05055. https://api.semanticscholar.org/CorpusID:266902774.

\bibitem[Kou et al.(2024)]{Kou2024EfficientBI}
Kou, K., Gao, X., Zhang, G., Xiong, Y., Nie, F., Bai, H., \& Gan, J. (2024). Efficient Blind Image Deblurring Network Based on Frequency Decomposition. \textit{IEEE Sensors Journal}, \textit{24}, 23212-23223. https://api.semanticscholar.org/CorpusID:270231056.

\bibitem[Ho et al.(2024)]{Ho2024EHNetEH}
Ho, Q.-T., Duong, M.-T., Lee, S., \& Hong, M.-C. (2024). EHNet: Efficient Hybrid Network with Dual Attention for Image Deblurring. \textit{Sensors (Basel, Switzerland)}, \textit{24}. https://api.semanticscholar.org/CorpusID:273319830.

\bibitem[Tanwar \& Susan(2024)]{r3}
Tanwar, M., \& Susan, S. (2024). Performance Evaluation of Deep Pre-Trained Models Under Progressive Blur. In \textit{2024 IEEE 16th International Conference on Computational Intelligence and Communication Networks (CICN)} (pp. 1254-1258). https://doi.org/10.1109/CICN63059.2024.10847407.

\bibitem[Zhang et al.(2019)]{r5}
Zhang, H., Dai, Y., Li, H., \& Koniusz, P. (2019). Deep Stacked Hierarchical Multi-Patch Network for Image Deblurring. In \textit{Proceedings of the IEEE/CVF Conference on Computer Vision and Pattern Recognition (CVPR)}. June.

\bibitem[Kupyn et al.(2019)]{r6}
Kupyn, O., Martyniuk, T., Wu, J., \& Wang, Z. (2019). DeblurGAN-v2: Deblurring (Orders-of-Magnitude) Faster and Better. In \textit{Proceedings of the IEEE/CVF International Conference on Computer Vision (ICCV)}. October.

\bibitem[Sun et al. (2025)]{r7}
Sun, R., Li, X., Zhang, L., Su, Y., Di, J., \& Liu, G. (2025). Wavelet-integrated deep neural network for deblurring and segmentation of crack images. \textit{Mechanical Systems and Signal Processing}, \textit{224}, 112240. https://doi.org/10.1016/j.ymssp.2024.112240.

\bibitem[Jiang et al. (2023)]{r8}
Jiang, N., Zhang, Y., Yan, F., Fu, X., \& Kong, T. (2023). Image blind motion deblurring method with longitudinal channel and wavelet dynamic convolution. \textit{Comput. Graph.}, \textit{116}, 275-286. https://doi.org/10.1016/j.cag.2023.08.022.

\bibitem[Ashar et al.(2024)]{r9}
Ashar, A. A. K., Abrar, A., \& Liu, J. (2024). A Survey on Object Detection and Recognition for Blurred and Low-Quality Images: Handling, Deblurring, and Reconstruction. In \textit{Proceedings of the 2024 8th International Conference on Information System and Data Mining} (pp. 95–100). https://doi.org/10.1145/3686397.3686413.

\bibitem[Che et al.(2024)]{r10}
Che, Y., Zheng, G., Li, Y., Hui, X., Li, Y. (2024). Unmanned Agricultural Machine Operation System in Farmland Based on Improved Fuzzy Adaptive Priority-Driven Control Algorithm. \textit{Electronics}, \textit{13}(20), 4141. https://doi.org/10.3390/electronics13204141.

\bibitem[Rattanpal et al. (2024)]{r11}
Rattanpal, S., Kashish, K., Kumari, T., Manvi, \& Gupta, S. (2024). Object Detection in Adverse Weather Conditions using Machine Learning. In \textit{2024 13th International Conference on System Modeling \& Advancement in Research Trends (SMART)} (pp. 239-247). https://doi.org/10.1109/SMART63812.2024.10882495.


\bibitem[Zhang et al. (2022)]{r12}
Zhang, K., Ren, W., Luo, W., et al. (2022). Deep Image Deblurring: A Survey. \textit{International Journal of Computer Vision}, \textit{130}(8), 2103--2130. https://doi.org/10.1007/s11263-022-01633-5.

\bibitem[Ma et al. (2023)]{r101}
Ma, W., Yu, H., Fang, W., Guan, F., Ma, D., Guo, Y., Zhang, Z., \& Wang, C. (2023). Crop Disease Detection against Complex Background Based on Improved Atrous Spatial Pyramid Pooling. \textit{Electronics}, \textit{12}(1), 216. https://doi.org/10.3390/electronics12010216.

\bibitem[Kong et al. (2022)]{r102}
Kong, J., Wang, H., Yang, C., Jin, X., Zuo, M., \& Zhang, X. (2022). A Spatial Feature-Enhanced Attention Neural Network with High-Order Pooling Representation for Application in Pest and Disease Recognition. \textit{Agriculture}, \textit{12}(4), 500. https://doi.org/10.3390/agriculture12040500.


\bibitem[Li et al. (2024)]{r103}
Li, Y., Li, Q., Pan, J., Zhou, Y., Zhu, H., Wei, H., \& Liu, C. (2024). SOD-YOLO: Small-Object-Detection Algorithm Based on Improved YOLOv8 for UAV Images. \textit{Remote Sensing}, \textit{16}(16), 3057. https://doi.org/10.3390/rs16163057.

\bibitem[Ma et al. (2024)]{r105}
Ma, H., Liu, Y., Xu, Z., \& Deng, W. (2024). Research on the DBG-YOLO Algorithm for Precise Identification of Corn Leaf Diseases in Farmland Environment. \textit{Journal of Hunan Agricultural University (Natural Sciences Edition)}.

\bibitem[Jing et al. (2024)]{r106}
Jing, R., Zhang, W., Liu, Y., Li, W., Li, Y., \& Liu, C. (2024). An effective method for small object detection in low-resolution images. \textit{Engineering Applications of Artificial Intelligence}, \textit{127}, 107206. https://doi.org/10.1016/j.engappai.2023.107206.

\bibitem[\protect\citeauthoryear{Zhang et al.}{{2023}}]{r107}
Zhang, J., Xiong, Y., Li, H., \& He, J. (2023). SDAN-YOLO: Self-attention domain adaptive network for YOLOv7. In \textit{2023 IEEE 29th International Conference on Parallel and Distributed Systems (ICPADS)} (pp. 420-427). https://doi.org/10.1109/ICPADS60453.2023.00070.

\bibitem[\protect\citeauthoryear{Bai et al.}{{2023}}]{r109}
Bai, D., Wang, S., Wang, W., Wang, H., Zhao, C., Yuan, P., \& Chen, Z. (2023). Overcoming Noisy Labels in Federated Learning Through Local Self-Guiding. In \textit{2023 IEEE/ACM 23rd International Symposium on Cluster, Cloud and Internet Computing (CCGrid)} (pp. 367-376). https://doi.org/10.1109/CCGrid57682.2023.00042.

\bibitem[\protect\citeauthoryear{Anwar et al.}{{2023}}]{r202}
Anwar, H., Khan, S. U., Ghaffar, M. M., Fayyaz, M., Khan, M. J., Weis, C., Wehn, N., \& Shafait, F. (2023). The NWRD Dataset: An Open-Source Annotated Segmentation Dataset of Diseased Wheat Crop. \textit{Sensors}, \textit{23}(15), 6942. https://doi.org/10.3390/s23156942.

\bibitem[\protect\citeauthoryear{Select Dataset}{{2025}}]{AppleGrowthVision}
Select Dataset. (2025). AppleGrowthVision | Precision agricultural data set | Computer Vision Dataset. Retrieved from https://www.selectdataset.com/dataset/fd9220b5ce834714b17b1fcdd7d445a0. 

\bibitem[\protect\citeauthoryear{Select Dataset}{{2025}}]{SemanticSugarBeets}
Select Dataset. (2025). SemanticSugarBeets. Retrieved from \url{https://www.selectdataset.com/dataset/3a07afafc87c5784d3f47d4d78d389c0.}

\bibitem[\protect\citeauthoryear{Mudavath \& Mamidi}{{2025}}]{r301}
Mudavath, T., \& Mamidi, A. (2025). Object detection challenges: Navigating through varied weather conditions—A comprehensive survey. \textit{Journal of Ambient Intelligence and Humanized Computing}, \textit{16}, 443--457. https://doi.org/10.1007/s12652-025-04956-6.

\bibitem[\protect\citeauthoryear{Munir et al.}{{2024}}]{r302}
Munir, A., Siddiqui, A. J., Anwar, S., El-Maleh, A., Khan, A. H., \& Rehman, A. (2024). Impact of Adverse Weather and Image Distortions on Vision-Based UAV Detection: A Performance Evaluation of Deep Learning Models. \textit{Drones}, \textit{8}(11), 638. https://doi.org/10.3390/drones8110638.

\bibitem[\protect\citeauthoryear{Lin et al.}{{2017}}]{Lin_2017_CVPR}
Lin, T.-Y., Dollar, P., Girshick, R., He, K., Hariharan, B., \& Belongie, S. (2017). Feature Pyramid Networks for Object Detection. In \textit{Proceedings of the IEEE Conference on Computer Vision and Pattern Recognition (CVPR)}. July.

\bibitem[\protect\citeauthoryear{Lin et al.}{{2023}}]{agriculture13030567}
Lin, S., Xiu, Y., Kong, J., Yang, C., \& Zhao, C. (2023). An Effective Pyramid Neural Network Based on Graph-Related Attentions Structure for Fine-Grained Disease and Pest Identification in Intelligent Agriculture. \textit{Agriculture}, \textit{13}(3), 567. https://doi.org/10.3390/agriculture13030567.

\bibitem[\protect\citeauthoryear{Cui et al.}{{2024}}]{agriculture10587008}
Cui, B., Liang, L., Ji, B., Zhang, L., Zhao, L., Zhang, K., Shi, F., \& Créput, J.-C. (2024). Exploring the YOLO-FT Deep Learning Algorithm for UAV-Based Smart Agriculture Detection in Communication Networks. \textit{IEEE Transactions on Network and Service Management}, \textit{21}(5), 5347-5360. https://doi.org/10.1109/TNSM.2024.3424232.

\bibitem[\protect\citeauthoryear{Park et al.}{{2023}}]{s23094432}
Park, H.-J., Kang, J.-W., \& Kim, B.-G. (2023). ssFPN: Scale Sequence (S2) Feature-Based Feature Pyramid Network for Object Detection. \textit{Sensors}, \textit{23}(9), 4432. https://doi.org/10.3390/s23094432.

\bibitem[\protect\citeauthoryear{Seifi \& Al-Mamun}{{2024}}]{Seifi2024}
Seifi, N., \& Al-Mamun, A. (2024). Optimizing Memory Access Efficiency in CUDA Kernel via Data Layout Technique. \textit{Journal of Computer and Communications}, \textit{12}, 124--139. https://doi.org/10.4236/jcc.2024.125009.

\bibitem[\protect\citeauthoryear{Ma et al.}{{2025}}]{ma2025nmspmmacceleratingmatrixmultiplication}
Ma, C., Wu, D., Deng, Z., Chen, J., Huang, X., Meng, J., Zhu, W., Zhou, A. C., Chen, P., Deng, M., Wei, Y., Feng, S., Pan, Y. (2025). NM-SpMM: Accelerating Matrix Multiplication Using N:M Sparsity with GPGPU. arXiv. https://arxiv.org/abs/2503.01253.

\bibitem[\protect\citeauthoryear{HoseinyFarahabady \& Zomaya}{{2025}}]{Out-of-Memory}
HoseinyFarahabady, M. R., \& Zomaya, A. Y. (2025). Out-of-Memory GPU Sorting Using Asynchronous CUDA Streams. In Y. Li, Y. Zhang, \& J. Xu (Eds.), \textit{Parallel and Distributed Computing, Applications and Technologies} (pp. 248-259). Springer Nature Singapore. https://doi.org/10.1007/978-981-96-4207-6\_23.

\bibitem[\protect\citeauthoryear{Zhang et al.}{{2020}}]{Zhang2020}
Zhang, H., Qian, Y., Wang, Y., et al. (2020). A ViBe Based Moving Targets Edge Detection Algorithm and Its Parallel Implementation. \textit{International Journal of Parallel Programming}, \textit{48}, 890--908. https://doi.org/10.1007/s10766-019-00628-z.

\bibitem[\protect\citeauthoryear{Shaikh et al.}{{2024}}]{doi:10.1142/S0219467824500190}
Shaikh, S. A., Chopade, J. J., \& Sardey, M. P. (2024). Real-Time Multi-Object Detection Using Enhanced Yolov5-7S on Multi-GPU for High-Resolution Video. \textit{International Journal of Image and Graphics}, \textit{24}(02), 2450019. https://doi.org/10.1142/S0219467824500190.

\bibitem[\protect\citeauthoryear{Menon et al.}{{2021}}]{9438307}
Menon, V. V., Siddiqui, S. A., Rao, S., Schmidt, A., French, M., Chirayath, V., \& Li, A. (2021). Design and Performance Evaluation of Multispectral Sensing Algorithms on CPU, GPU, and FPGA. In \textit{2021 IEEE Aerospace Conference (50100)} (pp. 1-9). https://doi.org/10.1109/AERO50100.2021.9438307.

\bibitem[\protect\citeauthoryear{Cui et al.}{{2024}}]{10587008}
Cui, B., Liang, L., Ji, B., Zhang, L., Zhao, L., Zhang, K., Shi, F., \& Créput, J.-C. (2024). Exploring the YOLO-FT Deep Learning Algorithm for UAV-Based Smart Agriculture Detection in Communication Networks. \textit{IEEE Transactions on Network and Service Management}, \textit{21}(5), 5347-5360. https://doi.org/10.1109/TNSM.2024.3424232.

\bibitem[\protect\citeauthoryear{Zhang et al.}{{2024}}]{Zhang_2024_08}
Zhang, Z., Zhang, P., Xu, Z., Yan, B., \& Wang, Q. (2024). Im2col-Winograd: An Efficient and Flexible Fused-Winograd Convolution for NHWC Format on GPUs. In \textit{Proceedings of the 53rd International Conference on Parallel Processing} (pp. 1072–1081). ACM. https://doi.org/10.1145/3673038.3673039.

\bibitem[\protect\citeauthoryear{Guo et al.}{{2024}}]{Guo_2024_09}
Guo, X., Yang, H., \& Jiang, H. (2024). Improved Medical Image Segmentation Method and Three-Dimensional Reconstruction Based on 3D-Unet. In \textit{2024 2nd International Conference on Signal Processing and Intelligent Computing (SPIC)} (pp. 881--885). IEEE. https://doi.org/10.1109/SPIC62469.2024.10691548.

\bibitem[\protect\citeauthoryear{ABE et al.}{{2008}}]{2008-67}
ABE, D., SEGAWA, E., NAKAYAMA, O., SHIOHARA, M., SASAKI, S., SUGANO, N., \& KANNO, H. (2008). Robust Small-Object Detection for Outdoor Wide-Area Surveillance. \textit{IEICE Transactions on Information and Systems}, \textit{E91.D}(7), 1922-1928. https://doi.org/10.1093/ietisy/e91-d.7.1922.

\bibitem[\protect\citeauthoryear{Nicholson}{{2023}}]{nicholson2023leveraging}
Nicholson, L. (2023). Leveraging deep-learned object detectors for real-world semantic localization and mapping. [PhD thesis, Queensland University of Technology].

\bibitem[\protect\citeauthoryear{Tan et al.}{2020}]{tan2020efficientdet} 
Tan, M., Pang, R. and Le, Q. V. (2020). Efficientdet: Scalable and efficient object detection. In \textit{Proceedings of the IEEE/CVF conference on computer vision and pattern recognition}, pp. 10781--10790. 

\bibitem[\protect\citeauthoryear{Xia}{2024}]{xia2024research} 
Xia, T. (2024). Research on Moving Object Real-time Recognition based on Deep Neural Network. In \textit{Proceedings of the International Conference on Machine Learning, Pattern Recognition and Automation Engineering}, pp. 199--204. 

\bibitem[\protect\citeauthoryear{Islam et al.}{2025}]{11130331} 
Islam, R., Mulé, J., Challagundla, D., Rizvi, S. and Carson, S. (2025). An Event Autoencoder for High-Speed Vision Sensing. In \textit{2025 IEEE Computer Society Annual Symposium on VLSI (ISVLSI)}, Vol.1, pp. 1--6. doi: \url{10.1109/ISVLSI65124.2025.11130331}. 

\bibitem[\protect\citeauthoryear{Liu and Kang}{2024}]{10757874} 
Liu, Y. and Kang, K.-D. (2024). AROD: Adaptive Real-Time Object Detection Based on Pixel Motion Speed. In \textit{2024 IEEE 100th Vehicular Technology Conference (VTC2024-Fall)}, pp. 1--7. doi: \url{10.1109/VTC2024-Fall63153.2024.10757874}. 

\bibitem[\protect\citeauthoryear{Feng et al.}{2023}]{feng4681618enhancing} FENG, H., GUO, J., XU, H., DU, Z., ZHANG, Y., HE, Y., CAO, G., GE, S. S. (2023). Enhancing Visual Surveillance on Smart Ships Through Integrated Image Deblurring and Object Detection. \textit{Available at SSRN 4681618}. 

\bibitem[\protect\citeauthoryear{Nandibewoor et al.}{2023}]{nandibewoor2023computer} NANDIBEWOOR, A., PRATEEK, L. K., SAKARAY, M., HASSAN, A., RAVANKAR, A., HEGDE, A. (2023). Computer Vision Application in Object Detection and Tracking for Aerial Surveillance. \textit{Indian Journal of Science and Technology}, \textit{16}(31), 2374--2379.

\bibitem[\protect\citeauthoryear{Yan}{2022}]{yan2022using} YAN, Y. (2022). Using the Improved SSD Algorithm to Motion Target Detection and Tracking. \textit{Computational Intelligence and Neuroscience}, \textit{2022}(1), 1886964. https://doi.org/10.1155/2022/1886964

\bibitem[\protect\citeauthoryear{Martins et al.}{2025}]{martins2025enhancing} MARTINS, F., BARAS, K., MARQUES, E. (2025). Enhancing Privacy: Using Blurred Images with MobileNet SSD. In \textit{EPIA Conference on Artificial Intelligence} (pp. 388--399). Springer.

\bibitem[\protect\citeauthoryear{Wu et al.}{2025}]{wu2025solidtrack} WU, D., HUANG, Z., ZHANG, Y. (2025). SolidTrack: A Novel Method for Robust and Reliable Multi-Pedestrian Tracking. \textit{Electronics}, \textit{14}(7), 1370. https://doi.org/10.3390/electronics14071370

\bibitem[\protect\citeauthoryear{Kiran et al.}{2025}]{kiran2025spatial}
KIRAN, D. G., KUMAR, T. R. (2025). Spatial Pyramid Pooling (SPP-NET) Compared with Convolutional Neural Networks to Recognise the Object from Given Images with Improved Accuracy. In \textit{AIP Conference Proceedings} (Vol. 3270, No. 1, p. 020058). AIP Publishing LLC. https://doi.org/10.1063/3270.020058


\bibitem[\protect\citeauthoryear{Zhang et al.}{2020}]{zhang2020insulator} ZHANG, X., ZHANG, Y., HU, M., JU, X. (2020). Insulator Defect Detection Based on YOLO and SPP-Net. In \textit{2020 International Conference on Big Data Artificial Intelligence Software Engineering (ICBASE)} (pp. 403--407). IEEE. https://doi.org/10.1109/ICBASE51474.2020.00085

\bibitem[\protect\citeauthoryear{Darnilasari et al.}{2023}]{10249515} DARNILASARI, A., INDRABAYU, ARENI, I. S. (2023). Implementation of Faster R-CNN with Colour and Blur Augmentation for Differentiate Cloves from Debris. In \textit{2023 2nd International Conference on Computer System, Information Technology, and Electrical Engineering (COSITE)} (pp. 72-77). IEEE. https://doi.org/10.1109/COSITE60233.2023.10249515

\bibitem[\protect\citeauthoryear{Wen et al.}{2021}]{wen2021video}
WEN, L., DING, J., \& LOFFELD, O. (2021). Video SAR Moving Target Detection Using Dual Faster R-CNN. \textit{IEEE Journal of Selected Topics in Applied Earth Observations and Remote Sensing}, \textit{14}, 2984--2994. https://doi.org/10.1109/JSTARS.2021.3062809

\bibitem[\protect\citeauthoryear{Sayed et al.}{2021}]{sayed2021improved} SAYED, M., BROSTOW, G. (2021). Improved Handling of Motion Blur in Online Object Detection. In \textit{Proceedings of the IEEE/CVF Conference on Computer Vision and Pattern Recognition} (pp. 1706--1716). IEEE/CVF. https://doi.org/10.1109/CVPR46437.2021.00175

\bibitem[\protect\citeauthoryear{Xu et al.}{2016}]{xu2016real} XU, L., LUO, H., HUI, B., CHANG, Z. (2016). Real-Time Robust Tracking for Motion Blur and Fast Motion via Correlation Filters. \textit{Sensors}, \textit{16}(9), 1443. https://doi.org/10.3390/s16091443


\bibitem[\protect\citeauthoryear{Yang et al.}{2018}]{yang2018application} YANG, K., GENG, F. (2018). Application of Faster R-CNN Model on Human Running Pattern Recognition. \textit{arXiv preprint arXiv:1811.05147}. https://doi.org/10.48550/arXiv.1811.05147

\bibitem[\protect\citeauthoryear{Zheng et al.}{2021}]{zheng2021deblur}
ZHENG, S., WU, Y., JIANG, S., LU, C., GUPTA, G. (2021). Deblur-YOLO: Real-Time Object Detection with Efficient Blind Motion Deblurring. In \textit{2021 International Joint Conference on Neural Networks (IJCNN)} (pp. 1–8). IEEE. https://doi.org/10.1109/IJCNN52387.2021.9534397

\bibitem[\protect\citeauthoryear{Olorunshola et al.}{2023}]{olorunshola2023comparative} OLORUNSHOLA, O., JEMITOLA, P., ADEMUWAGUN, A. (2023). Comparative Study of Some Deep Learning Object Detection Algorithms: R-CNN, Fast R-CNN, Faster R-CNN, SSD, and YOLO. \textit{Nile Journal of Engineering and Applied Sciences}, \textit{1}(1), 70–80.

\bibitem[\protect\citeauthoryear{Okur and Kilicarslan}{2024}]{okur2024two} OKUR, O., KILICARSLAN, M. (2024). Two-Stream YOLOv8: Object and Motion Detection in Driving Videos. \textit{IEEE Transactions on Intelligent Vehicles}, \textit{1}(1), 1-10.

\end{thebibliography}
\end{document}